\documentclass[final,12pt]{elsarticle}

\usepackage{graphicx}
\usepackage{amssymb}
\usepackage{amsmath}

\usepackage[utf8]{inputenc}
\usepackage[T1]{fontenc}
\usepackage{todonotes}
\usepackage{url}
\usepackage{tabularx}
\usepackage{booktabs}
\usepackage[capitalize]{cleveref}
\usepackage{subcaption}

\newcolumntype{Y}{>{\centering\arraybackslash}X}

\journal{ISPRS Journal}

\begin{document}

\begin{frontmatter}


\title{Beyond RGB: Very High Resolution Urban Remote Sensing With Multimodal Deep Networks}

\author{Nicolas Audebert\corref{cor1}\fnref{label1,label2}}
\ead{nicolas.audebert@onera.fr}
\author{Bertrand Le Saux\fnref{label1}}
\ead{bertrand.le\_saux@onera.fr}
\author{Sébastien Lefèvre\fnref{label2}}
\ead{sebastien.lefevre@irisa.fr}

\address[label1]{ONERA, \textit{The French Aerospace Lab}, F-91761 Palaiseau, France}
\address[label2]{Univ. Bretagne-Sud, UMR 6074, IRISA, F-56000 Vannes, France}

\begin{abstract}
In this work, we investigate various methods to deal with semantic labeling of very high resolution multi-modal remote sensing data. Especially, we study how deep fully convolutional networks can be adapted to deal with multi-modal and multi-scale remote sensing data for semantic labeling. Our contributions are three-fold: a) we present an efficient multi-scale approach to leverage both a large spatial context and the high resolution data, b) we investigate early and late fusion of Lidar and multispectral data, c) we validate our methods on two public datasets with state-of-the-art results. Our results indicate that late fusion make it possible to recover errors steaming from ambiguous data, while early fusion allows for better joint-feature learning but at the cost of higher sensitivity to missing data.

\end{abstract}


\end{frontmatter}


\section{Introduction}
\label{S:intro}

Remote sensing has benefited a lot from deep learning in the past few years, mainly thanks to progress achieved in the computer vision community on natural RGB images. Indeed, most deep learning architectures designed for multimedia vision can be used on remote sensing optical images. This resulted in significant improvements in many remote sensing tasks such as vehicle detection~\cite{chen_vehicle_2014}, semantic labeling~\cite{audebert_semantic_2016,marmanis_semantic_2016,maggiori_fully_2016,sherrah_fully_2016} and land cover/use classification~\cite{penatti_deep_2015,nogueira_towards_2016}. However, these improvements have been mostly restrained to traditional 3-channels RGB images that are the main focus of the computer vision community.

On the other hand, Earth Observation data is rarely limited to this kind of optical sensor. Additional data, either from the same sensor (e.g. multispectral data) or from another one (e.g. a Lidar point cloud) is sometimes available. However, adapting vision-based deep networks to these larger data is not trivial, as this requires to work with new data structures that do not share the same underlying physical and numerical properties. Nonetheless, all these sources provide complementary information that should be used jointly to maximize the labeling accuracy.

In this work we present how to build a comprehensive deep learning model to leverage multi-modal high-resolution remote sensing data, with the example of semantic labeling of Lidar and multispectral data over urban areas. Our contributions are the following:
\begin{itemize}
	\item We show how to implement an efficient multi-scale deep fully convolutional neural network using SegNet~\cite{badrinarayanan_segnet:_2015} and ResNet~\cite{he_deep_2016}.
    \item We investigate early fusion of multi-modal remote sensing data based on the FuseNet principle~\cite{hazirbas_fusenet:_2016}. We show that while early fusion significantly improves semantic segmentation by allowing the network to learn jointly stronger multi-modal features, it also induces higher sensitivity to missing or noisy data.
    \item We investigate late fusion of multi-modal remote sensing data based on the residual correction strategy~\cite{audebert_semantic_2016}. We show that, although performing not as good as early fusion, residual correction improves semantic labeling and makes it possible to recover some critical errors on hard pixels.
    \item We successfully validate our methods on the ISPRS Semantic Labeling Challenge datasets of Vaihingen and Potsdam~\cite{cramer_dgpf_2010}, with results placing our methods amongst the best of the state-of-the-art.
\end{itemize}

\section{Related Work}
\label{S:related}
Semantic labeling of remote sensing data relates to the dense pixel-wise classification of images, which is called either ``semantic segmentation'' or ``scene understanding'' in the computer vision community. Deep learning has proved itself to be both effective and popular on this task, especially since the introduction of Fully Convolutional Networks (FCN)~\cite{long_fully_2015}. By replacing standard fully connected layers of traditional Convolutional Neural Networks (CNN) by convolutional layers, it was possible to densify the single-vector output of the CNN to achieve a dense classification at 1:8 resolution. The first FCN model has quickly been improved and declined in several variants. Some improvements have been based on convolutional auto-encoders with a symetrical architecture such as SegNet~\cite{badrinarayanan_segnet:_2015} and DeconvNet~\cite{noh_learning_2015}. Both use a bottleneck architecture in which the feature maps are upsampled to match the original input resolution, therefore performing pixel-wise predictions at 1:1 resolution. These models have however been outperformed on multimedia images by more sophisticated approaches, such as removing the pooling layers from standard CNN and using dilated convolutions~\cite{yu_multi-scale_2015} to preserve most of the input spatial information, which resulted in models such as the multi-scale DeepLab~\cite{chen_deeplab:_2016} which performs predictions at several resolutions using separate branches and produces 1:8 predictions. Finally, the rise of the residual networks~\cite{he_deep_2016} was soon followed by new architectures derived from ResNet~\cite{Pohlen_2017_CVPR,Zhao_2017_CVPR}. These architectures leverage the state-of-the-art effectiveness of residual learning for image classification by adapting them for semantic segmentation, again at a 1:8 resolution. All these architectures were shown to perform especially well on popular semantic segmentation of natural images benchmarks such as Pascal VOC~\cite{everingham_pascal_2014} and COCO~\cite{lin_microsoft_2014}.

On the other hand, deep learning has also been investigated for multi-modal data processing. Using dual-stream autoencoders,~\cite{ngiam_multimodal_2011} successfully jointly processed audio-video data using an architecture with two branches, one for audio and one for video that merge in the middle of the network. Moreover, processing RGB-D (or 2.5D) data has a significant interest for the computer vision and robotics communities, as many embedded sensors can sense both optical and depth information. Relevant architectures include two parallel networks CNN merging in the same fully connected layers~\cite{eitel_multimodal_2015} (for RGB-D data classification) and two CNN streams merging in the middle~\cite{guo_two-stream_2016} (for fingertip detection). FuseNet~\cite{hazirbas_fusenet:_2016} extended this idea to fully convolutional networks for semantic segmentation of RGB-D data by integrating an early fusion scheme into the SegNet architecture.  Finally, the recent work of~\cite{Park_2017_ICCV} builds on the FuseNet architecture to incorporate residual learning and multiple stages of refinement to obtain high resolution multi-modal predictions RGB-D data. These models can be used to learn jointly from several heterogeneous data sources;, although they focus on multimedia images.

As deep learning significantly improved computer vision tasks, remote sensing adopted those techniques and deep networks have been often used for Earth Observation. Since the first successful use of patch-based CNN for roads and buildings extraction~\cite{mnih_learning_2010}, many models were built upon the deep learning pipeline to process remote sensing data. For example,~\cite{saito_multiple_2016} performed multiple label prediction (i.e. both roads and buildings) in a single CNN.~\cite{vakalopoulou_building_2015} extended the approach to multispectral images including visible and infrared bands. Although successful, the patch-based classification approach only produces coarse maps, as an entire patch gets associated with only one label. Dense maps can be obtained by sliding a window over the entire input, but this is an expensive and slow process.
Therefore, for urban scenes with dense labeling in very high resolution, superpixel-based classification~\cite{campos-taberner_processing_2016} of urban remote sensing images was a successful approach that classified homogeneous regions to produce dense maps, as it combines the patch-based approach with an unsupervised pre-segmentation. Thanks to concatenating features fed to the SVM classifier,~\cite{audebert_how_2016,lagrange_benchmarking_2015} managed to extend this framework to multi-scale processing using a superpixel-based pyramidal approach.
Other approaches for semantic segmentation included patch-based prediction with mixed deep and expert features~\cite{paisitkriangkrai_effective_2015}, that used prior knowledge and feature engineering to improve the deep network predictions. Multi-scale CNN predictions have been investigated by \cite{liu_learning_2016}~with a pyramid of images used as input to an ensemble of CNN for land cover use classification, while \cite{chen_vehicle_2014} used several convolutional blocks to process multiple scales.
Lately, semantic labeling of aerial images has moved to FCN models~\cite{sherrah_fully_2016,maggiori_fully_2016,volpi_dense_2016}.
Indeed, Fully Convolutional Networks  such as SegNet or DeconvNet directly perform pixel-wise classification are very well suited for semantic mapping of Earth Observation data, as they can capture the spatial dependencies between classes without the need for pre-processing such as a superpixel segmentation, and they produce high resolution predictions. These approaches have again been extended for sophisticated multi-scale processing in \cite{marmanis_classification_2016}~using both the expensive pyramidal approach with an FCN and the multiple resolutions output inspired from~\cite{chen_deeplab:_2016}. Multiple scales allow the model to capture spatial relationships for objects of different sizes, from large arrangements of buildings to individual trees, allowing for a better understanding of the scene.
To enforce a better spatial regularity, probabilistic graphical models such as Conditional Random Fields (CRF) post-processing have been used to model relationships between neighboring pixels and integrate these priors in the prediction~\cite{lin_efficient_2015,sherrah_fully_2016,Liu_2017_CVPR_Workshops}, although this add expensive computations that significantly slow the inference. On the other hand, \cite{marmanis_classification_2016} proposed a network that learnt both the semantic labeling and the explicit inter-class boundaries to improve the spatial structure of the predictions. However, these explicit spatial regularization schemes are expensive. In this work, we aim to show that these are not necessary to obtain semantic labeling results that are competitive with the state-of-the-art.

Previously, works investigated fusion of multi-modal data for remote sensing. Indeed, complementary sensors can be used on the same scene to measure several properties that give different insights on the semantics of the scene. Therefore, data fusion strategies can help obtain better models that can use these multiple data modalities. To this end,~\cite{paisitkriangkrai_effective_2015} fused optical and Lidar data by concatenating deep and expert features as inputs to random forests. Similarly, \cite{Liu_2017_CVPR_Workshops} integrates expert features from the ancillary data (Lidar and NDVI) into their higher-order CRF to improve the main optical classification network. The work of \cite{audebert_semantic_2016} investigated late fusion of Lidar and optical data for semantic segmentation using prediction fusion that required no feature engineering by combining two classifiers with a deep learning end-to-end approach. This was also investigated in~\cite{Audebert_2017_CVPR_Workshops} to fuse optical and OpenStreetMap for semantic labeling. During the Data Fusion Contest (DFC) 2015,~\cite{lagrange_benchmarking_2015} proposed an early fusion scheme of Lidar and optical data based on a stack of deep features for superpixel-based classification of urban remote sensed data. In the DFC~2016, \cite{mou_spatiotemporal_2016} performed land cover classification and traffic analysis by fusing multispectral and video data at a late stage. Our goal is to thoroughly study end-to-end deep learning approaches for multi-modal data fusion and to compare early and late fusion strategies for this task. 

\section{Method description}
\label{S:method}

\subsection{Semantic segmentation of aerial images}

\begin{figure}[t]
	\includegraphics[width=\textwidth]{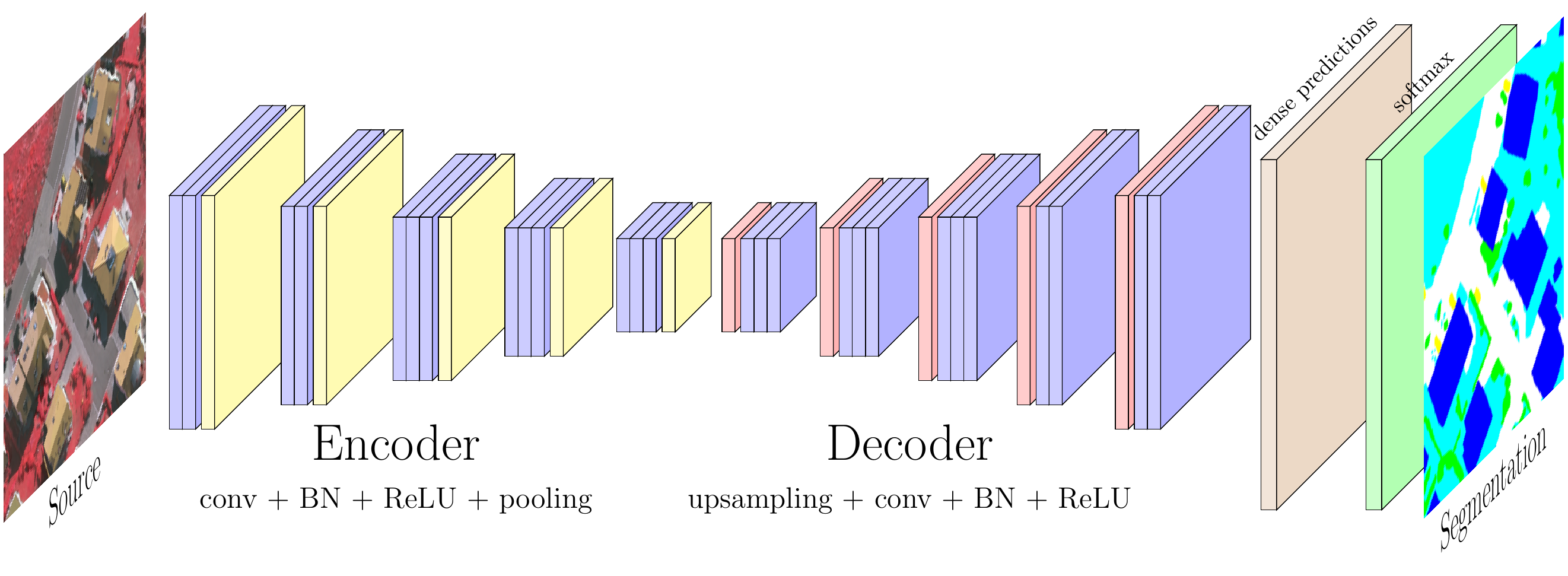}
    \caption{SegNet architecture~\cite{badrinarayanan_segnet:_2015} for semantic labeling of remote sensing data. See text for more detailed explanations of each layer.}
    \label{fig:segnet}
\end{figure}

Semantic labeling of aerial image requires a dense pixel-wise classification of the images. Therefore, we can use FCN architectures to achieve this, using the same techniques that are effective for natural images. We choose the SegNet~\cite{badrinarayanan_segnet:_2015} model as the base network in this paper. SegNet is based on an encoder-decoder architecture that produces an output with the same resolution as the input, as illustrated in~\cref{fig:segnet}. This is a desirable property as we want to label the data at original image resolution, therefore producing maps at 1:1 resolution compared to the input. SegNet allows such task to do as the decoder is able to upsample the feature maps using the unpooling operation. We also compare this base network to a modified version of the ResNet-34 network~\cite{he_deep_2016} adapted for semantic segmentation.

The encoder from SegNet is based on the convolutional layers from VGG-16~\cite{simonyan_very_2014}. It has 5 convolution blocks, each containing 2 or 3 convolutional layers of kernel $3\times3$ with a padding of 1 followed by a rectified linear unit (ReLU) and a batch normalization (BN)~\cite{ioffe_batch_2015}. Each convolution block is followed by a max-pooling layer of size $2\times2$. Therefore, at the end of the encoder, the feature maps are each $\frac{W}{32}\times\frac{H}{32}$ where the original image has a resolution $W\times H$.

\begin{figure}[t]
	\includegraphics[width=\textwidth]{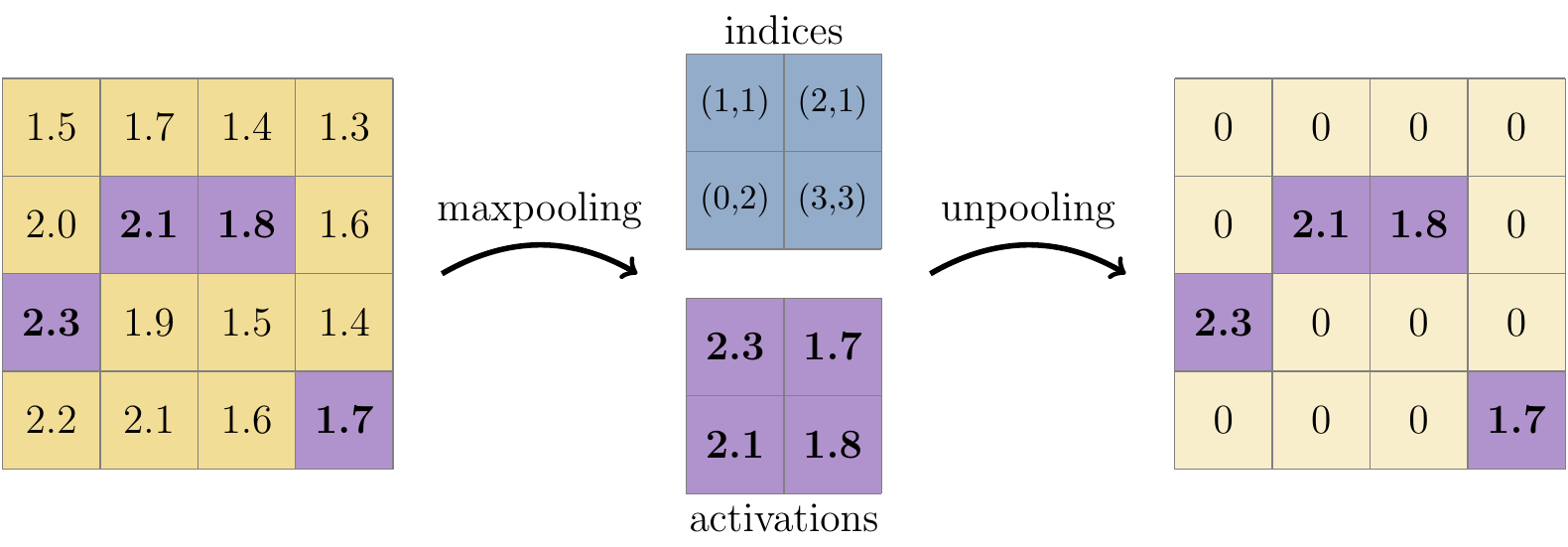}
	\caption{Illustration of the effects of the maxpooling and unpooling operations on a $4\times4$ feature map.}
    \label{fig:unpooling}
\end{figure}

The decoder performs both the upsampling and the classification. It learns how to restore the full spatial resolution while transforming the encoded feature maps into the final labels. Its structure is symmetrical with respect to the encoder. Pooling layers are replaced by unpooling layers as described in~\cite{zeiler_visualizing_2014}. The unpooling relocates the activation from the smaller feature maps into a zero-padded upsampled map. The activations are relocated at the indices computed at the pooling stages, i.e. the $argmax$ from the max-pooling (cf.~\cref{fig:unpooling}). This unpooling allows to replace the highly-abstracted features of the decoder to the saliency points of the low-level geometrical feature maps of the encoder. This is especially effective on small objects that would otherwise be misplaced or misclassified. After the unpooling, the convolution blocks densify the sparse feature maps. This process is repeated until the feature maps reach the input resolution. 

According to~\cite{he_deep_2016}, residual learning helps train deeper networks and achieved new state-of-the-art classification performance on ImageNet, as well as state-of-the-art semantic segmentation results on the COCO dataset. Consequently, we also compare our methods applied to the ResNet-34 architecture. ResNet-34 model uses four residual blocks. Each block is comprised of 2 or 3 convolutions of $3\times3$ kernels and the input of the block is summed into the output using a skip connection. As in SegNet, convolutions are followed by Batch Normalization and ReLU activation layers. The skip connection can be either the identity if the tensor shapes match, or a $1\times1$ convolution that projects the input feature maps into the same space as the output ones if the number of convolution planes changed. In our case, to keep most of the spatial resolution, we keep the initial $2\times2$ max-pooling but reduce the stride of all convolutions to 1. Therefore, the output of the ResNet-34 model is a 1:2 prediction map. To upsample this map back to full resolution, we perform an unpooling followed by a standard convolutional block.

Finally, both networks use a softmax layer to compute the multinomial logistic loss, averaged over the whole patch:
\begin{equation}
loss = \frac{1}{N} \sum_{i=1}^N \sum_{j=1}^k y_j^i \log\left(\frac{\exp(z_j^i)}{\sum\limits_{l=1}^k \exp(z_l^i)}\right)~,
\end{equation}
where $N$ is the number of pixels in the input image, $k$ the number of classes and, for a specified pixel $i$, $y^i$ denote its label and $(z^i_1,\dots, z^i_k)$ the prediction vector. This means that we only minimize the average pixel-wise classification loss without any spatial regularization, as it will be learnt by the network during training. We do not use any post-processing, \textit{e.g.} a CRF, as it would significantly slow down the computations for little to no gain.

\subsection{Multi-scale aspects}

\begin{figure}
	\begin{subfigure}{0.48\textwidth}
		\includegraphics[width=\textwidth]{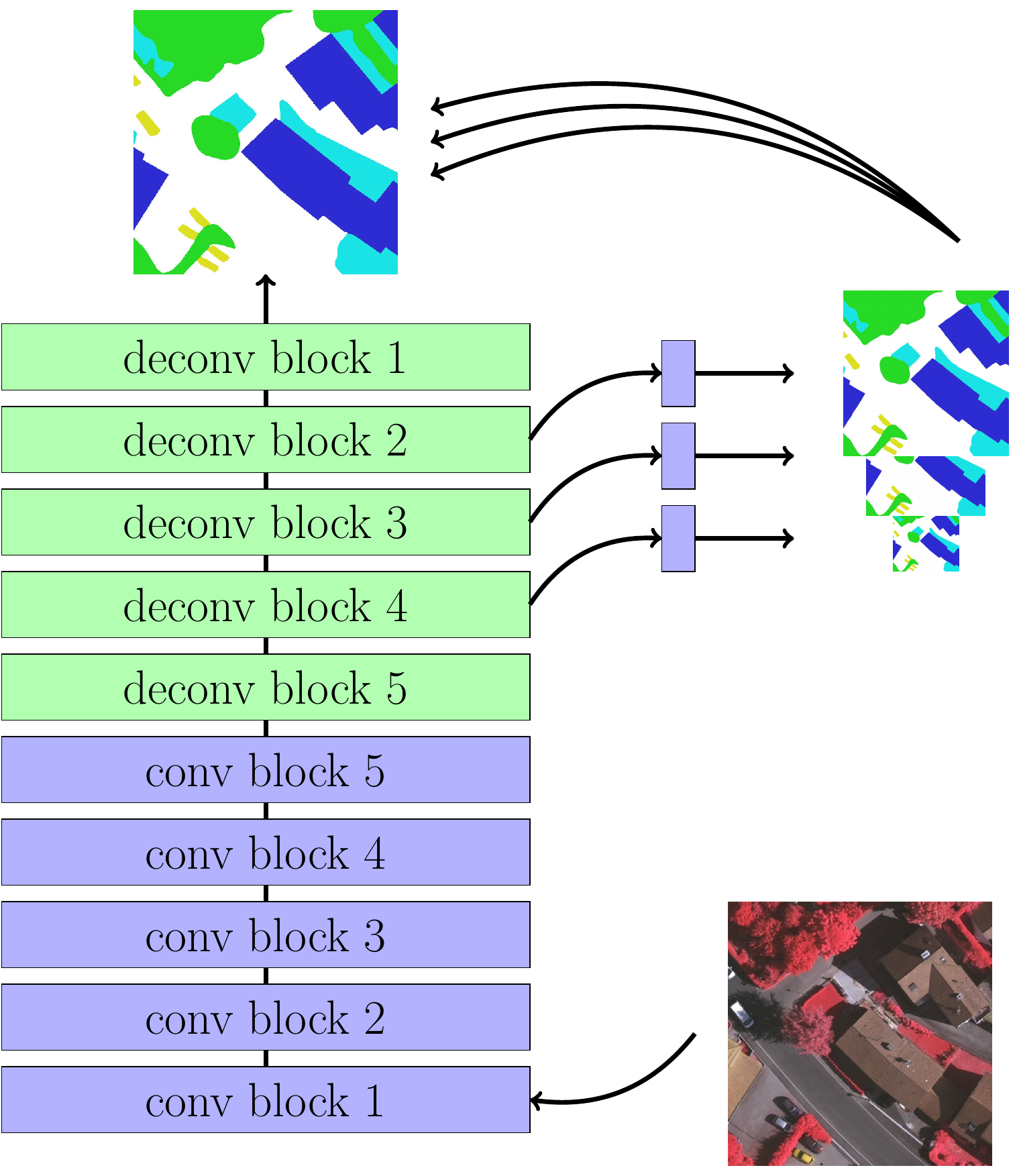}
        \caption{Multi-scale prediction using SegNet.}
    \end{subfigure}
    \begin{subfigure}{0.48\textwidth}
		\includegraphics[width=\textwidth]{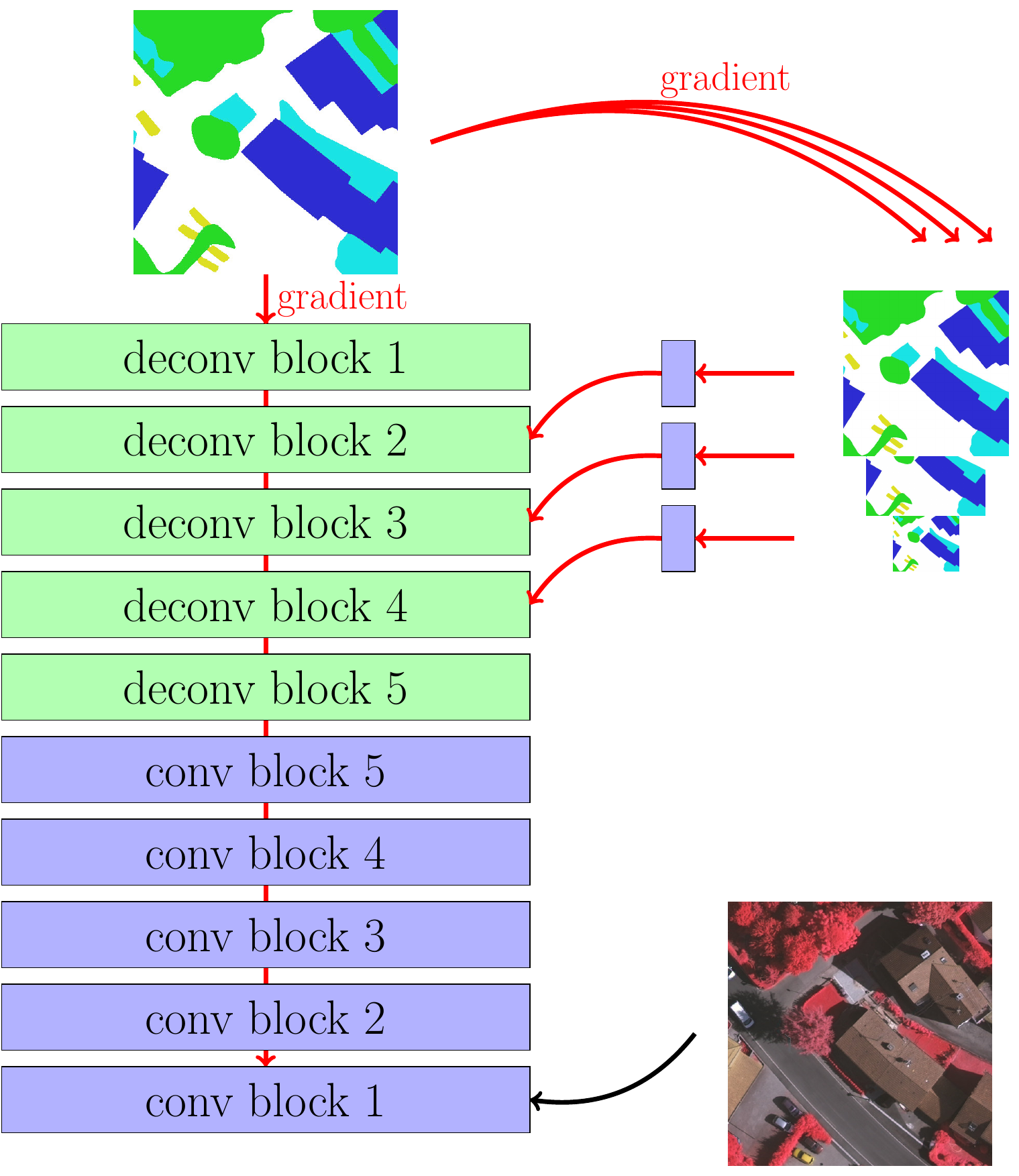}
        \caption{Backpropagation at multiple scales.}
    \end{subfigure}
    \caption{Multi-scale deep supervision of SegNet with 3 branches on remote sensing data.}
    \label{fig:ms_deep_segnet}
\end{figure}

Often multi-scale processing is addressed using a pyramidal approach: different context sizes and different resolutions are fed as parallel inputs to one or multiple classifiers. Our first contribution is the study of an alternative approach which consists in branching our deep network to generate output predictions at several resolutions. Each output has its own loss which is backpropagated to earlier layers of the network, in the same way as when performing deep supervision~\cite{lee_deeply-supervised_2014}. This is the approach that has been used for the DeepLab~\cite{chen_deeplab:_2016} architecture.

Therefore, considering our SegNet model, we not only predict one semantic map at full resolution, but we also branch the model earlier in the decoder to predict maps of smaller resolutions. After the $p$\textsuperscript{th} convolutional block of the decoder, we add a convolution layer that projects the feature maps into the label space, with a resolution $\frac{2^p W}{32} \times \frac{2^p H}{32}$, as illustrated in~\cref{fig:ms_deep_segnet}. Those smaller maps are then interpolated to full resolution and averaged to obtain the final full resolution semantic map.

Let $P_{full}$ denote the full resolution prediction, $P_{down_d}$ the predictions at the downscale factor $d$ and $f_d$ the bilinear interpolation that upsamples a map by a factor $d$. Therefore, we can aggregate our multi-resolution predictions using a simple summation (with $f_0 = Id$), \textit{e.g.} if we use four scales:
\begin{equation}
P_{full} = \sum_{d \in \{0, 2, 4, 8\}} f_d(P_{down_d}) = P_0 + f_2(P_2) + f_4(P_4) + f_8(P_8).
\end{equation}

During backpropagation, each branch will receive two contributions:
\begin{itemize}
	\item The contribution coming from the loss of the average prediction.
    \item The contribution coming from its own downscaled loss.
\end{itemize}
This ensures that earlier layers still have a meaningful gradient, even when the global optimization is converging. As argued in~\cite{lin_refinenet:_2016}, deeper layers now only have to learn how to refine the coarser predictions from the lower resolutions, which helps the overall learning process.

\subsection{Early fusion}

\begin{figure}
	\begin{subfigure}{\textwidth}
    \includegraphics[width=\textwidth]{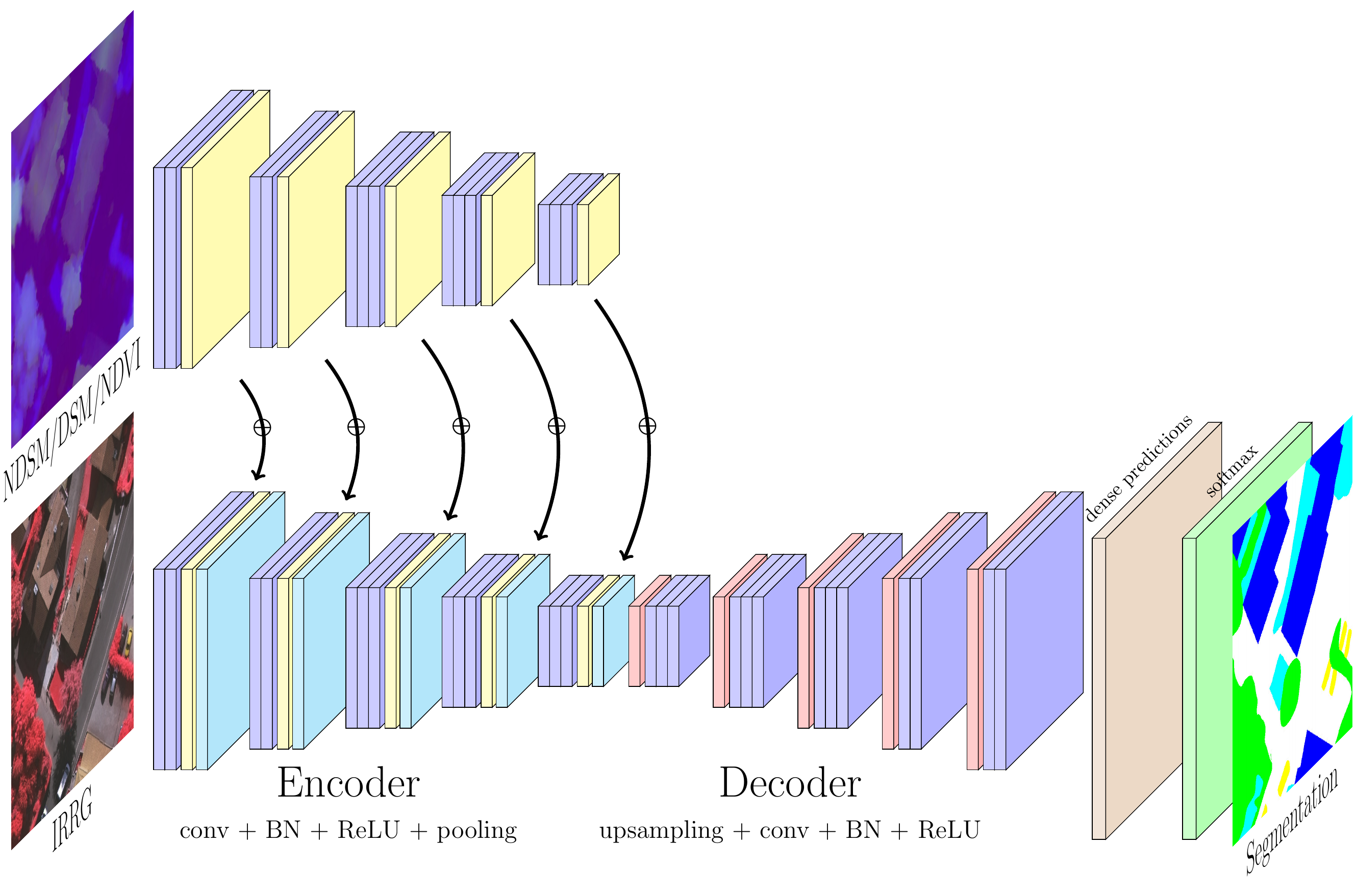}
    \caption{FuseNet architecture~\cite{hazirbas_fusenet:_2016} for early fusion of remote sensing data.}
    \label{fig:fusenet}
    \end{subfigure}
    \begin{subfigure}{\textwidth}
    \includegraphics[width=\textwidth]{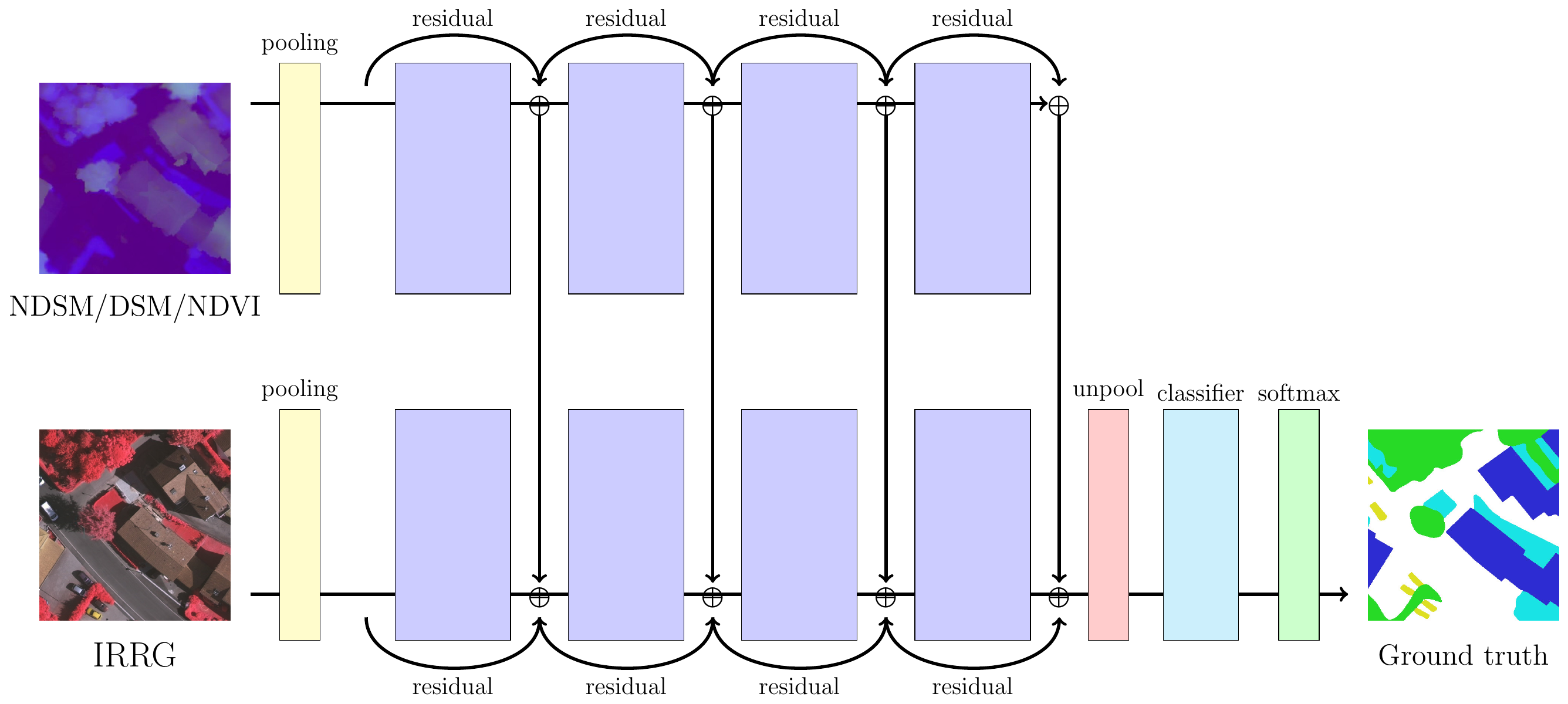}
    \caption{FusResNet : the FuseNet architecture adapted to a residual network.}
    \label{fig:fusresnet}
    \end{subfigure}
    \caption{Architectures of altered baselines FCN to fit the FuseNet framework.}
\end{figure}

\begin{figure}
	\begin{subfigure}{0.48\textwidth}
    	\includegraphics[width=\textwidth]{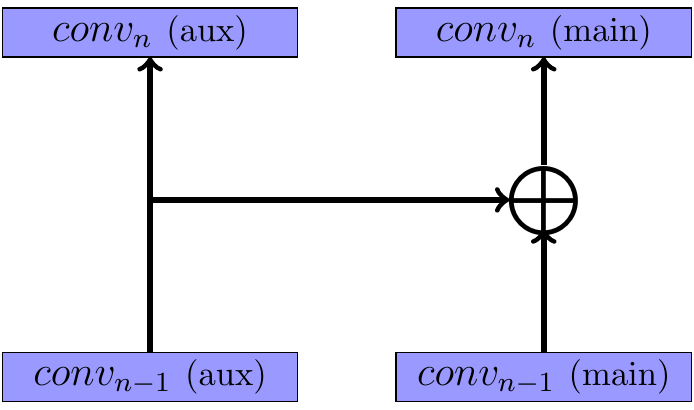}
        \caption{Original FuseNet: fuses contributions by summing auxiliary activations into the main branch.}
        \label{fig:fusenet_sum}
    \end{subfigure}
    \hfill
    \begin{subfigure}{0.48\textwidth}
    	\includegraphics[width=\textwidth]{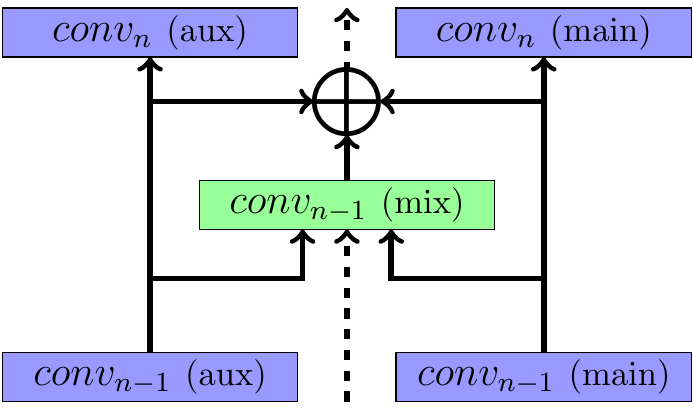}
        \caption{Our FuseNet: fuses contributions with a convolutional block followed by summation.}
        \label{fig:fusenet_mix}
    \end{subfigure}
    \caption{Fusion strategies for the FuseNet architecture.}
\end{figure}

In the computer vision community, RGB-D images are often called 2.5D images. Integrating this data into deep learning models has proved itself to be quite challenging as the naive stacking approach does not perform well in practice. Several data fusion schemes have been proposed to work around this obstacle. The FuseNet~\cite{hazirbas_fusenet:_2016} approach uses the SegNet architecture in a multi-modal context. As illustrated in~\cref{fig:fusenet}, it jointly encodes both the RGB and depth information using two encoders whose contributions are summed after each convolutional block. Then, a single decoder upsamples the encoded joint-representation back into the label probability space. This data fusion approach can also be adapted to other deep neural networks, such as residual networks as illustrated in~\cref{fig:fusresnet}.

However, in this architecture the depth data is treated as second-hand. Indeed, the two branches are not exactly symmetrical: the depth branch works only with depth-related information where as the optical branch actually deals with a mix of depth and optical data. Moreover, in the upsampling process, only the indices from the main branch will be used. Therefore, one needs to choose which data source will be the primary one and which one will be the auxiliary data~(cf.~\cref{fig:fusenet_sum}). There is a conceptual unbalance in the way the two sources are dealt with. We suggest an alternative architecture with a third ``virtual'' branch that does not have this unbalance, which might improve performance.

Instead of computing the sum of the two sets of feature maps, we suggest an alternative fusion process to obtain the multi-modal joint-features. We introduce a third encoder that does not correspond to any real modality, but instead to a virtual fused data source. At stage $n$, the virtual encoder takes as input its previous activations concatenated with both activations from the other encoders. These feature maps are passed through a convolutional block to learn a residual that is summed with the average feature maps from the other encoders. This is illustrated in~\cref{fig:fusenet_mix}. This strategy makes FuseNet symetrical and therefore relieves us of the choice of the main source, which would be an additional hyperparameter to tune. This architecture will be named \textbf{V-FuseNet} in the rest of the paper for Virtual-FuseNet.

\subsection{Late fusion}

\begin{figure}[!htb]
	\begin{subfigure}{\textwidth}
	\includegraphics[width=\textwidth]{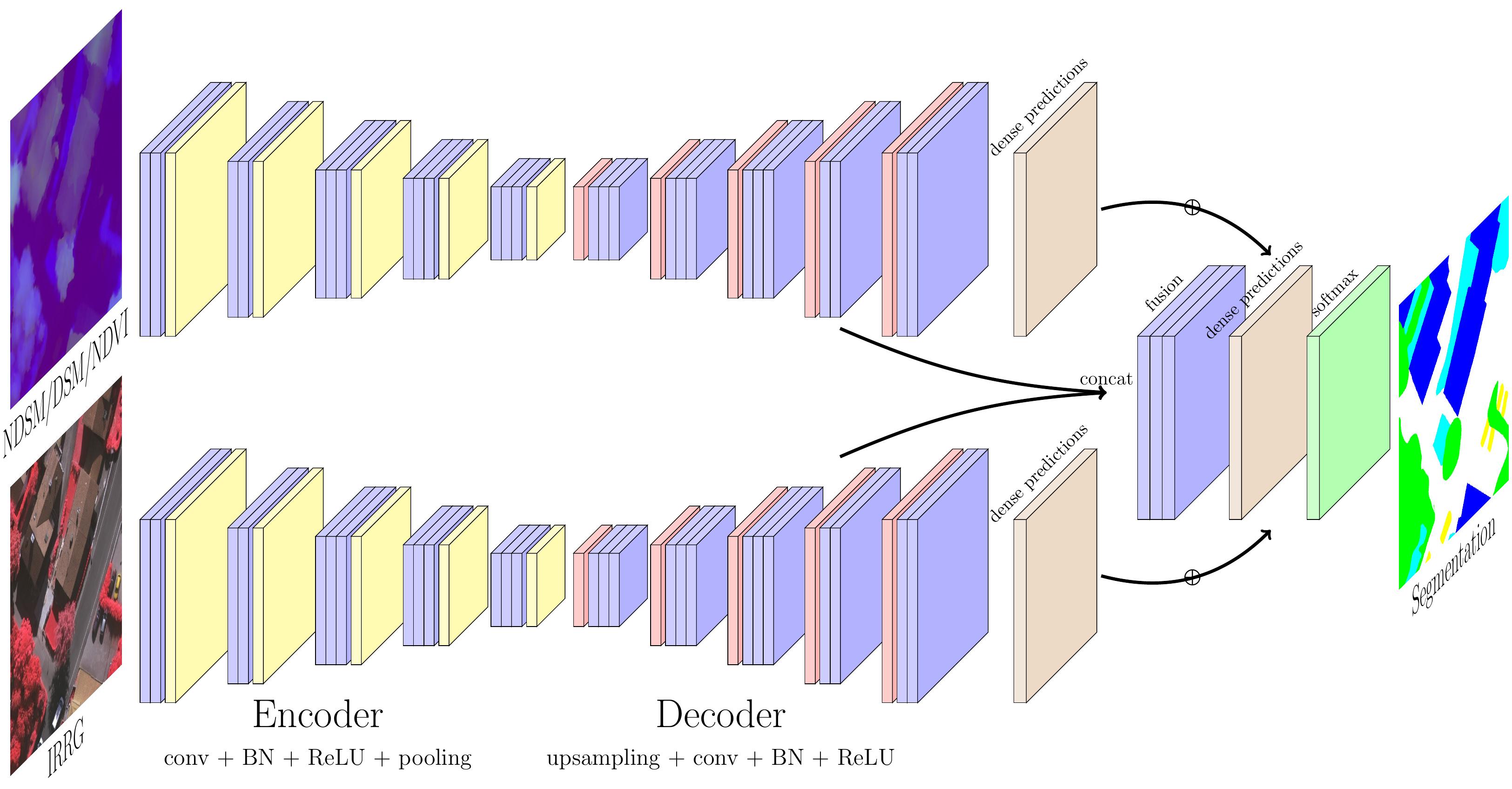}
    \caption{Residual correction~\cite{audebert_semantic_2016} for late fusion using two SegNets.}
    \label{fig:residual_correction}
    \end{subfigure}
    \begin{subfigure}{\textwidth}
    \includegraphics[width=\textwidth]{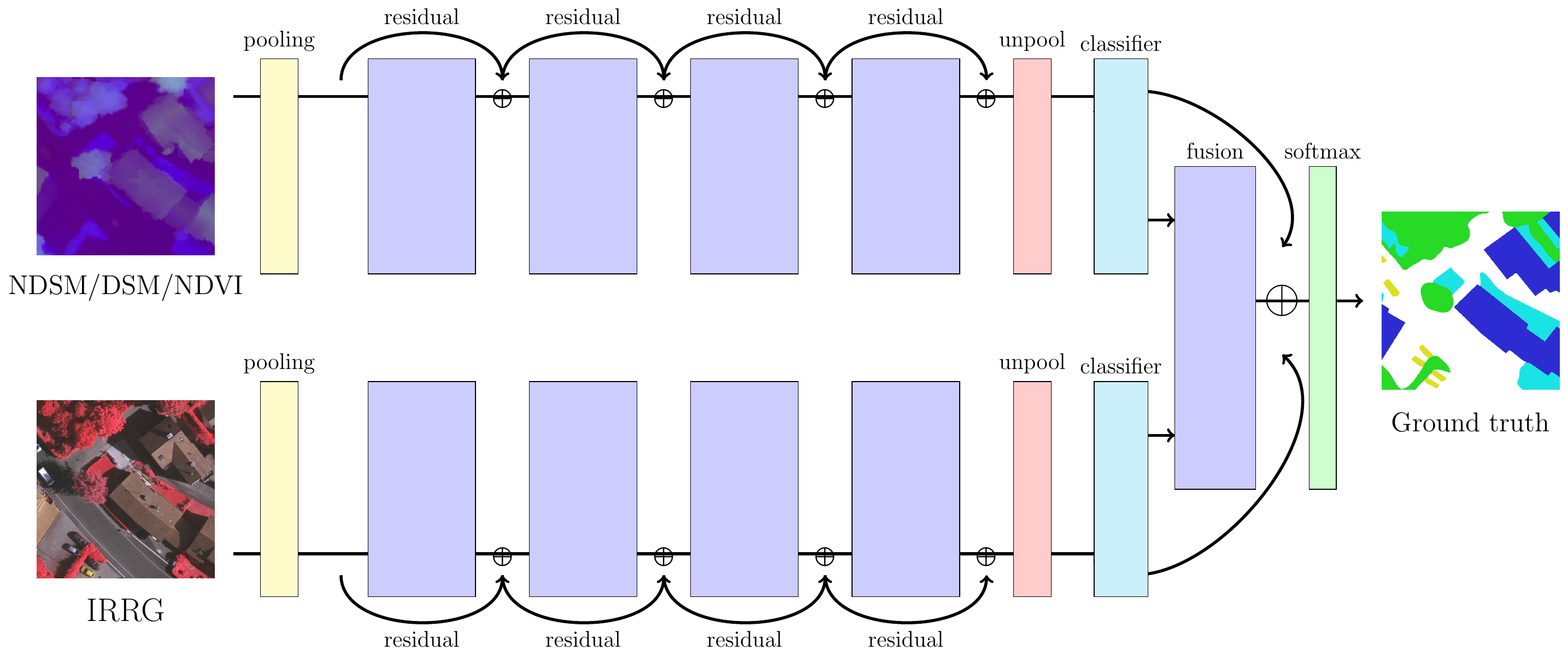}
    \caption{Residual correction~\cite{audebert_semantic_2016} for late fusion using two ResNets.}
    \label{fig:residual_correction_resnet}
    \end{subfigure}
    \caption{Architectures of altered baselines FCN to fit the residual correction framework.}
\end{figure}

One caveat of the FuseNet approach is that both streams are expected to be topologically compatible in order to fuse the encoders. However, this might not always be the case, especially when dealing with data that does not possess the same structure (e.g. 2D images and a 3D point cloud). Therefore, we propose an alternative fusion technique that relies only on the late feature maps with no assumption on the models. Specifically, instead of investigating fusion at the data level, we work around the data heterogeneity by trying to achieve prediction fusion. This process was investigated in~\cite{audebert_semantic_2016} where a residual correction module was introduced. This module consists in a residual convolutional neural network that takes as input the last feature maps from two deep networks. Those networks can be topologically identical or not. In our case, each deep network is a Fully Convolutional Network that has been trained on either the optical or the auxiliary data source. Each FCN generates a prediction. First, we average the two predictions to obtain a smooth classification map. Then, we re-train the correction module in a residual fashion. The residual correction network therefore learns a small offset to apply to each pixel-probabilities. This is illustrated in~\cref{fig:residual_correction} for the SegNet architecture and~\cref{fig:residual_correction_resnet} for the ResNet architecture.

Let $R$ be the number of outputs on which to perform residual correction, $P_0$ the ground truth, $P_i$ the prediction and $\epsilon_i$ the error term from $P_i$ w.r.t. the ground truth. We predict $P'$, the sum of the averaged predictions and the correction term $c$ which is inferred by the fusion network:
\begin{equation}
P' = P_{avg} + c = \frac{1}{R} \sum_{i=1}^R P_i + c = P_0 + \frac{1}{R} \sum_{i=1}^R \epsilon_i + c~~,
\end{equation}

As our residual correction module is optimized to minimize the loss, we enforce:
\begin{equation}
\lVert P' - P_0 \rVert \rightarrow 0
\end{equation}
which translates into a constraint on $c$ and $\epsilon_i$:
\begin{equation}
\lVert \frac{1}{R} \sum_{i=1}^R \epsilon_i - c \rVert \rightarrow 0~.
\end{equation}

As this offset $c$ is learnt in a supervised way, the network can infer which input to trust depending on the predicted classes. For example, if the auxiliary data is better for vegetation detection, the residual correction will attribute more weight to the prediction coming out of the auxiliary SegNet. This module can be generalized to $n$ inputs, even with different network architectures. This architecture will be denoted \textbf{SegNet-RC} (for SegNet-Residual Correction) in the rest of the paper.

\subsection{Class balancing}

The remote sensing datasets (later described in~\cref{S:datasets}) we consider have unbalanced semantic classes. Indeed, the relevant structures in urban areas do not occupy the same surface (\textit{i.e.} the same number of pixels) in the images. Therefore, when performing semantic segmentation, the class frequencies can be very inhomogeneous. To improve the class average accuracy, we balance the loss using the inverse class frequencies. However, as one of the considered class is a reject class (``clutter'') that is also very rare, we do not use inverse class frequency for this one. Instead, we apply the same weight on this class as the lowest weight on all the other classes. This takes into account that the clutter class is an ill-posed problem anyway.

\section{Experiments}
\label{S:results}

\begin{table}
    \caption{Validation results on Vaihingen.}
    \label{table:val_vaihingen}
	\begin{tabularx}{\textwidth}{Y c c}
    \toprule
    Model & Overall accuracy & Average F1\\
    \midrule
    SegNet (IRRG) & 90.2 {\small $\pm$ 1.4} & 89.3 {\small $\pm$ 1.2}\\
    SegNet (composite) & 88.3 {\small $\pm$ 0.9} & 81.6 {\small $\pm$ 0.8}\\
    SegNet-RC & 90.6 {\small $\pm$ 1.4} & 89.2 {\small $\pm$ 1.2}\\
    FuseNet & 90.8 {\small $\pm$ 1.4} & 90.1 {\small $\pm$ 1.2}\\
    V-FuseNet & \textbf{91.1} {\small $\pm$ 1.5} & \textbf{90.3} {\small $\pm$ 1.2}\\
    \midrule
    ResNet-34 (IRRG) & 90.3 {\small $\pm$ 1.0} & 89.1 {\small $\pm$ 0.7}\\
    ResNet-34 (composite) & 88.8 {\small $\pm$ 1.1} & 83.4 {\small $\pm$ 1.3}\\
    ResNet-34-RC & 90.8 {\small $\pm$ 1.0} & 89.1 {\small $\pm$ 1.1}\\
    FusResNet & 90.6 {\small $\pm$ 1.1} & 89.3 {\small $\pm$ 0.7}\\
    \bottomrule
    \end{tabularx}
\end{table}


\begin{table}
    \caption{Multi-scale results on Vaihingen.}
    \label{table:msc_vaihingen}
	\begin{tabularx}{\textwidth}{Y c c c c c c}
    \toprule
    Number of branches & imp. surf. & buildings & low veg. & trees & cars & Overall\\
    \midrule
    No branch & 92.2 & 95.5 & \textbf{82.6} & \textbf{88.1} & 88.2 & 90.2 {\small $\pm$ 1.4}\\
    1 branch & 92.4 & 95.7 & 82.3 & 87.9 & \textbf{88.5} & 90.3 {\small $\pm$ 1.5}\\
    2 branches & 92.5 & \textbf{95.8} & 82.4 & 87.8 & 87.6 & 90.3 {\small $\pm$ 1.4}\\
    3 branches & \textbf{92.7} & \textbf{95.8} & \textbf{82.6} & \textbf{88.1} & 88.1 & \textbf{90.5} {\small $\pm$ 1.5}\\
    \bottomrule
    \end{tabularx}
\end{table}

\begin{table}
    \caption{Final results on the Vaihingen dataset.}
    \label{table:final_vaihingen}
    \setlength\tabcolsep{5pt}
	\begin{tabularx}{\textwidth}{Y c c c c c c}
    \toprule
	Method & imp. surf. & buildings & low veg. & trees & cars & Overall\\
    \midrule
    FCN~\cite{sherrah_fully_2016} & 90.5 & 93.7 & 83.4 & 89.2 & 72.6 & 89.1\\
    FCN + fusion + boundaries~\cite{marmanis_classification_2016} & \textbf{92.3} & \textbf{95.2} & 84.1 & \textbf{90.0} & 79.3 & \textbf{90.3}\\
    \midrule
    SegNet (IRRG) & 91.5 & 94.3 & 82.7 & 89.3 & 85.7 & 89.4\\
	SegNet-RC & 91.0 & 94.5 & 84.4 & 89.9 & 77.8 & 89.8\\
    FuseNet & 91.3 & 94.3 & \textbf{84.8} & 89.9 & 85.9 & 90.1\\
    V-FuseNet & 91.0 & 94.4 & 84.5 & 89.9 & \textbf{86.3} & 90.0\\
    \bottomrule
    \end{tabularx}
\end{table}

\begin{table}
    \caption{Final results on the Potsdam dataset.}
    \label{table:final_potsdam}
    \setlength\tabcolsep{4pt}
	\begin{tabularx}{\textwidth}{Y c c c c c c}
    \toprule
	Method & imp. surf. & buildings & low veg. & trees & cars & Overall\\
    \midrule
    FCN + CRF + expert features~\cite{Liu_2017_CVPR_Workshops} & 91.2 & 94.6 & 85.1 & 85.1 & 92.8 & 88.4\\
    FCN~\cite{sherrah_fully_2016} & 92.5 & \textbf{96.4} & 86.7 & 88.0 & 94.7 & 90.3\\
    \midrule
    SegNet (IRRG) & 92.4 & 95.8 & 86.7 & 87.4 & 95.1 & 90.0\\
	SegNet-RC & 91.3 & 95.9 & 86.2 & 85.6 & 94.8 & 89.0\\
    V-FuseNet & \textbf{92.7} & 96.3 & \textbf{87.3} & \textbf{88.5} & \textbf{95.4} & \textbf{90.6}\\
    \bottomrule
    \end{tabularx}
\end{table}

\subsection{Datasets}
\label{S:datasets}
We validate our method on the two image sets of the ISPRS 2D Semantic Labeling Challenge \footnote{\url{http://www2.isprs.org/commissions/comm3/wg4/semantic-labeling.html}}. These datasets are comprised of very high resolution aerial images over two cities in Germany: Vaihingen and Potsdam. The goal is to perform semantic labeling of the images on six classes : buildings, impervious surfaces (\textit{e.g.} roads), low vegetation, trees, cars and clutter. Two online leaderboards (one for each city) are available and report test metrics obtained on held-out test images.

\paragraph{ISPRS Vaihingen}
The Vaihingen dataset has a resolution of 9~cm/pixel with tiles of approximately $2100\times2100$ pixels. There are 33 images, from which 16 have a public ground truth. Tiles consist in Infrared-Red-Green (IRRG) images and DSM data extracted from the Lidar point cloud. We also use the normalized DSM (nDSM) from~\cite{gerke_use_2015}.

\paragraph{ISPRS Potsdam}
The Potsdam dataset has a resolution of 5~cm/pixel with tiles of $6000\times6000$. There are 38 images, from which 24 have a public ground truth. Tiles consist in Infrared-Red-Green-Blue (IRRGB) multispectral images and DSM data extracted from the Lidar point cloud. nDSM are also included in the dataset with two different methods.

\subsection{Experimental setup}
For each optical image, we compute the NDVI using the following formula:
\begin{equation}
NDVI = \frac{IR - R}{IR + R}~.
\end{equation}
We then build a composite image comprised of the stacked DSM, nDSM and NDVI.

As the tiles are very high resolution, we cannot process them directly in our deep networks. We use a sliding window approach to extract $128\times128$ patches. The stride of the sliding window also defines the size of the overlapping regions between two consecutive patches. At training time, a smaller stride allows us to extract more training samples and acts as data augmentation. At testing time, a smaller stride allows us to average predictions on the overlapping regions, which reduces border effects and improves the overall accuracy. During training, we use a $64$px stride for Potsdam and a $32$px stride for Vaihingen. We use a $32$px stride for testing on Potsdam and a $16$px stride on Vaihingen.

Models are implemented using the Caffe framework. We train all our models using Stochastic Gradient Descent (SGD) with a base learning rate of 0.01, a momentum of 0.9, a weight decay of 0.0005 and a batch size of 10. For SegNet-based architectures, the weights of the encoder in SegNet are initialized with those of VGG-16 trained on ImageNet, while the decoder weights are randomly initalized using the policy from~\cite{he_delving_2015}. We divide the learning rate by 10 after 5, 10 and 15 epochs. For ResNet-based models, the four convolutional blocks are initialized using weights from ResNet-34 trained on ImageNet, the other weights being initialized using the same policy. We divide the learning rate by 10 after 20 and 40 epochs. In both cases, the learning rate of the pre-initialized weights is set as half the learning of the new weights as suggested in~\cite{audebert_semantic_2016}.

Results are cross-validated on each dataset using a 3-fold split. Final models for testing on the held-out data are re-trained on the whole training set.

\subsection{Results}

\begin{figure}[!htb]
	\hfill
	\begin{subfigure}{0.45\textwidth}
    \includegraphics[width=\linewidth]{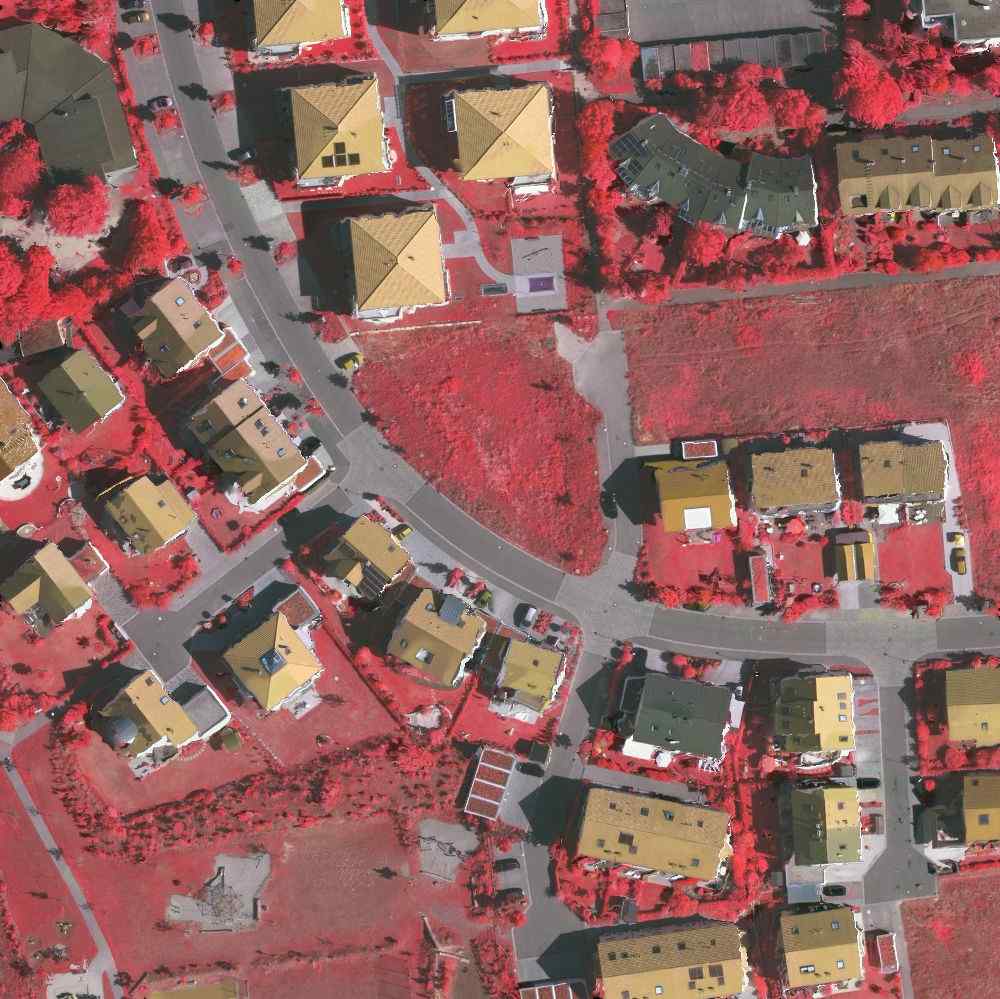}
    \caption{IRRG image}
    \end{subfigure}
    \hfill
    \begin{subfigure}{0.45\textwidth}
    \includegraphics[width=\linewidth]{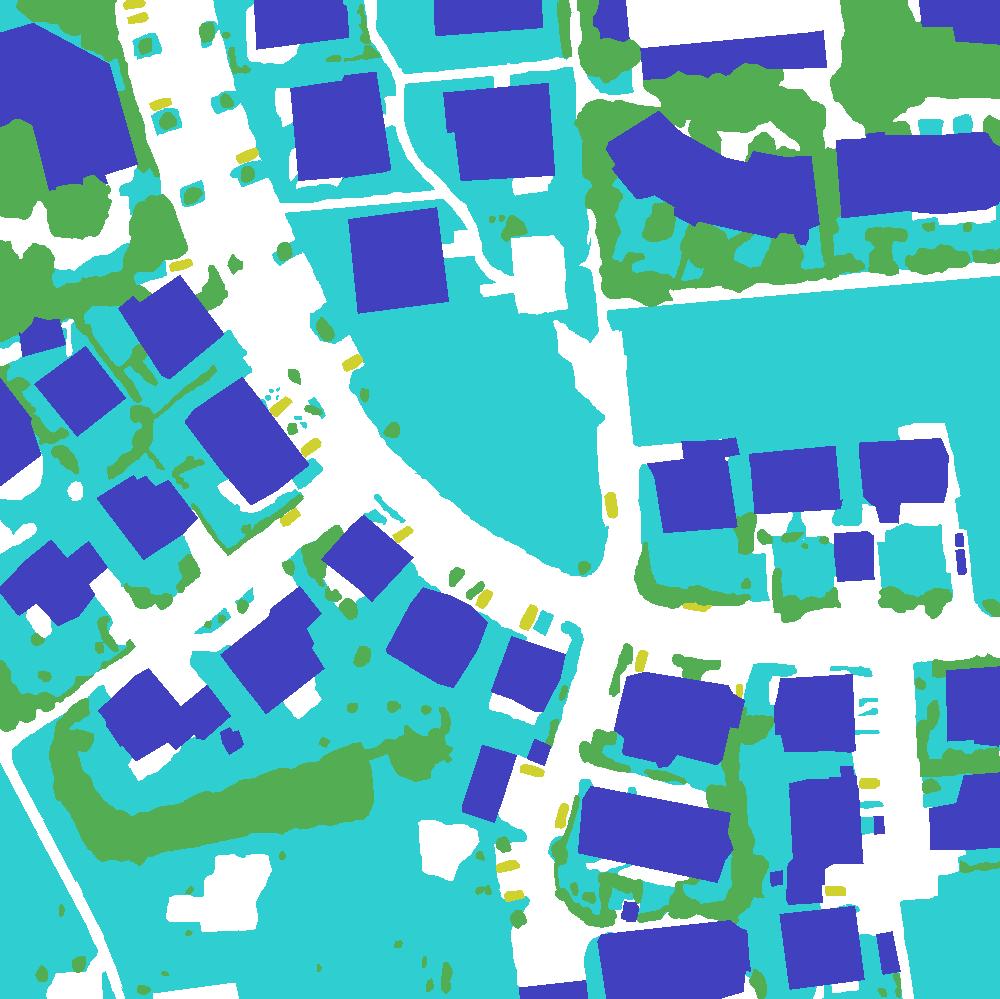}
    \caption{Ground truth}
    \end{subfigure}
    \hfill
    
    \hfill
    \begin{subfigure}{0.45\textwidth}
    \includegraphics[width=\linewidth]{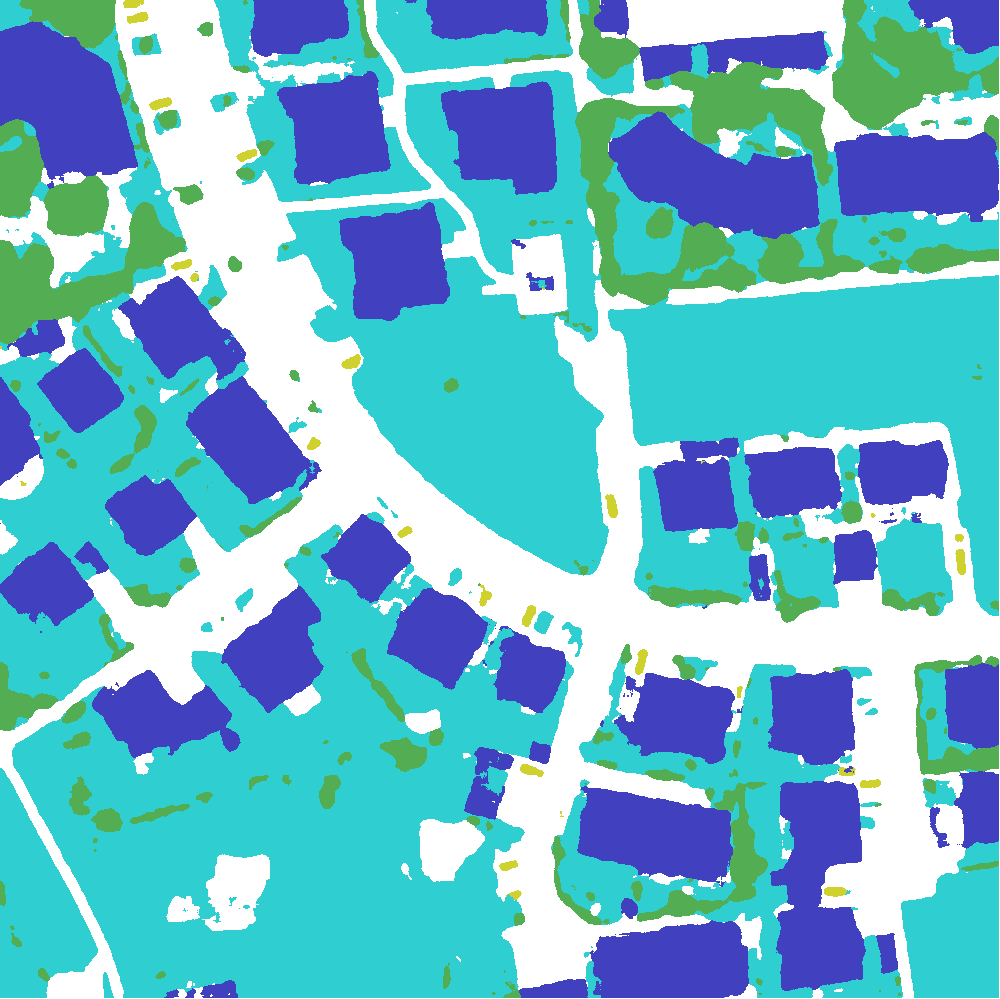}
    \caption{SegNet standard}
    \end{subfigure}
    \hfill
    \begin{subfigure}{0.45\textwidth}
    \includegraphics[width=\linewidth]{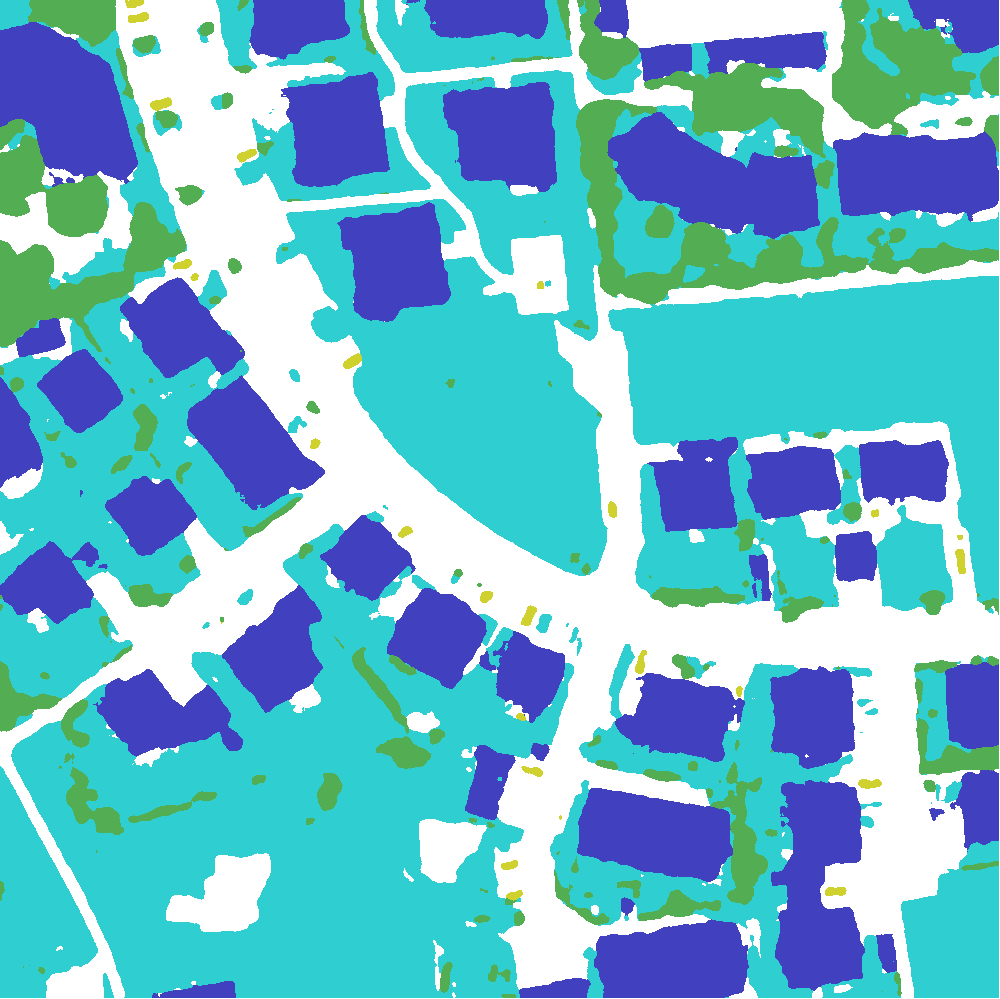}
    \caption{SegNet multi-scale (3 scales)}
    \end{subfigure}
    \hfill
    \caption{Effect of the multi-scale prediction strategy on a excerpt of the ISPRS Vaihingen dataset. Small objects or surfaces with ambiguous spatial context are regularized by the multiple scales prediction aggregation.\\
(white: roads, {\color{blue!80!black} blue}: buildings, {\color{cyan!80!black} cyan}: low vegetation, {\color{green!80!black} green}:~trees, {\color{yellow!80!black} yellow}: cars)}
    \label{fig:vaihingen_images}
\end{figure}

\begin{figure}[!htb]
	\begin{subfigure}{0.32\textwidth}
    \includegraphics[width=\linewidth]{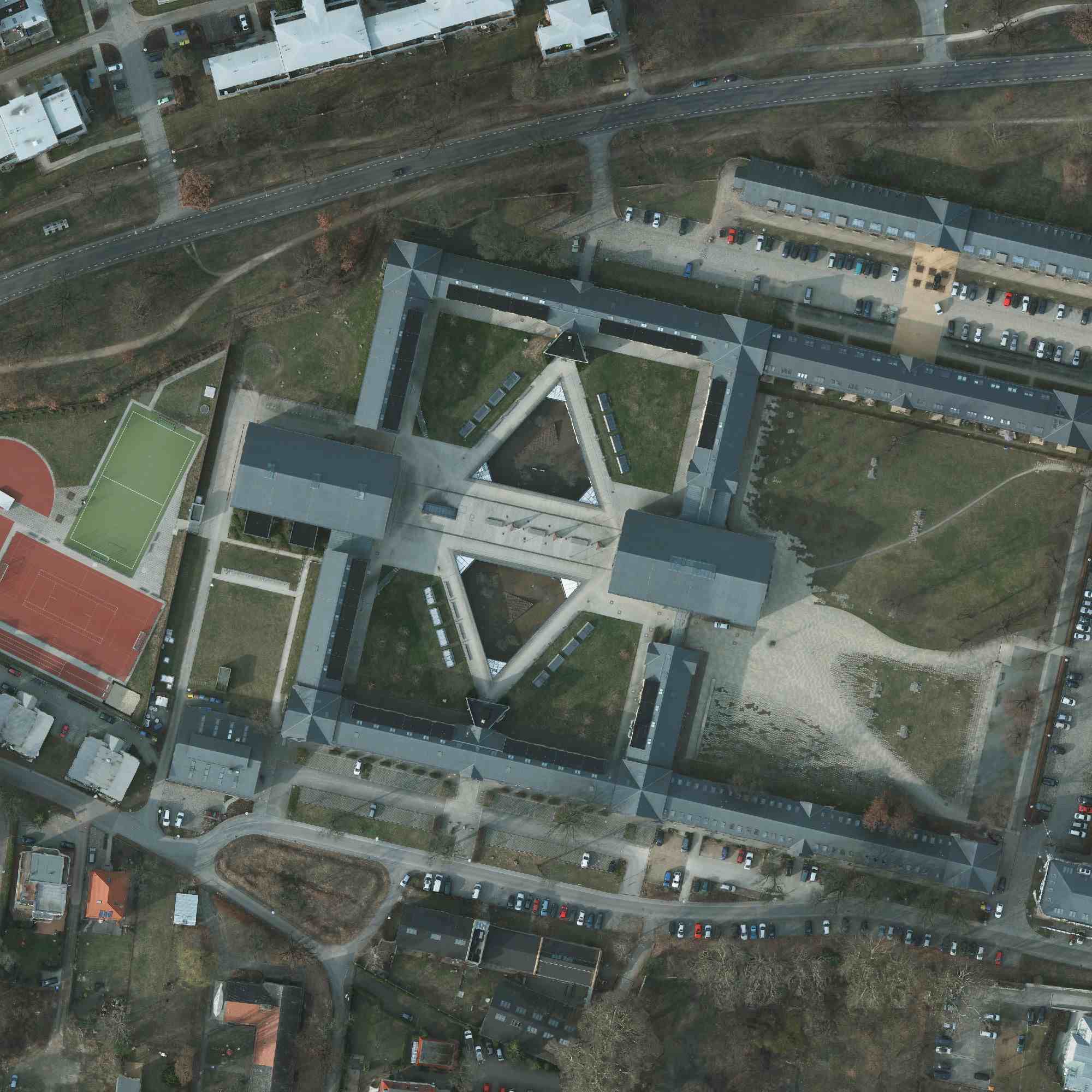}
    \caption{RGB image}
    \end{subfigure}
    \begin{subfigure}{0.32\textwidth}
    \includegraphics[width=\linewidth]{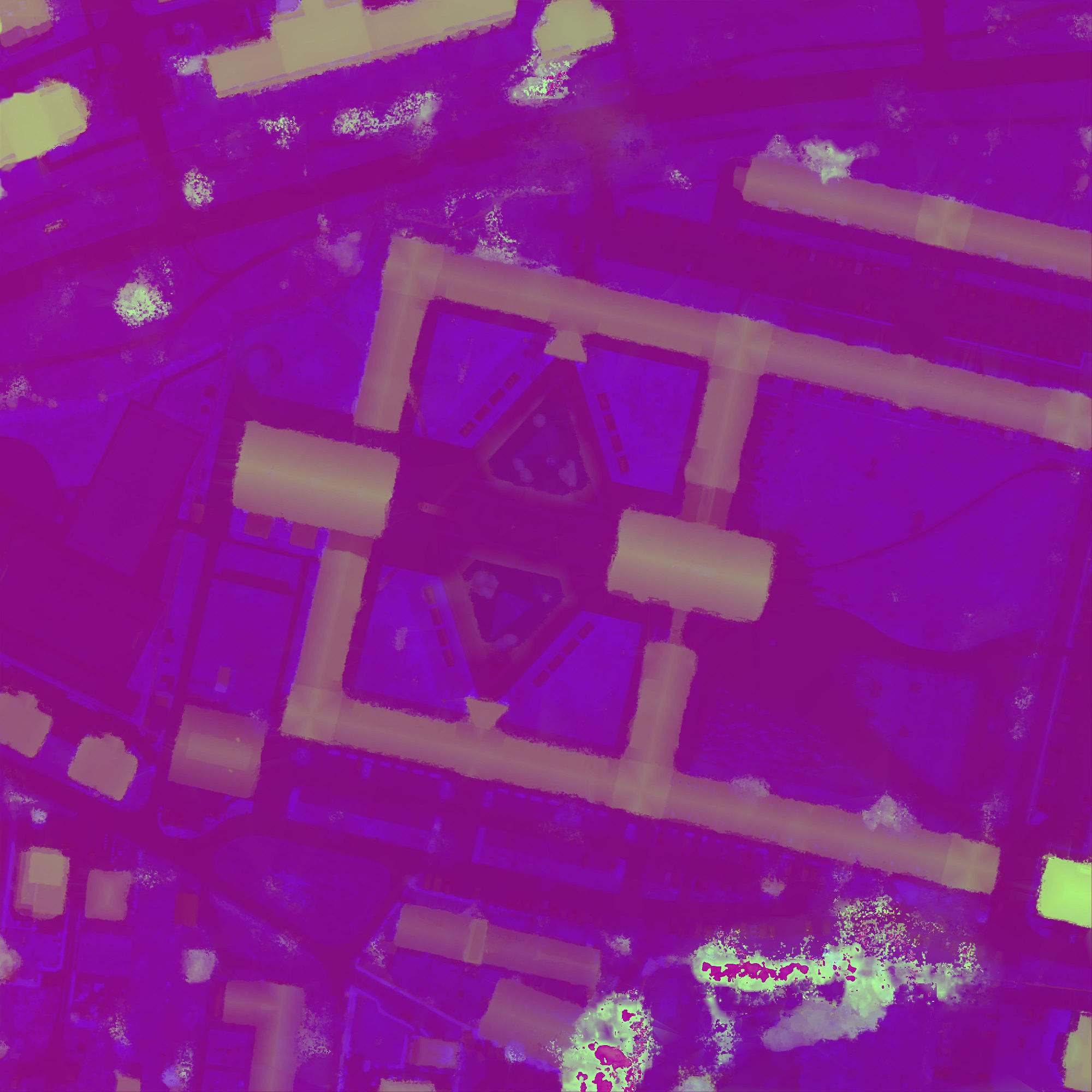}
    \caption{Composite image}
    \end{subfigure}
    \begin{subfigure}{0.32\textwidth}
    \includegraphics[width=\linewidth]{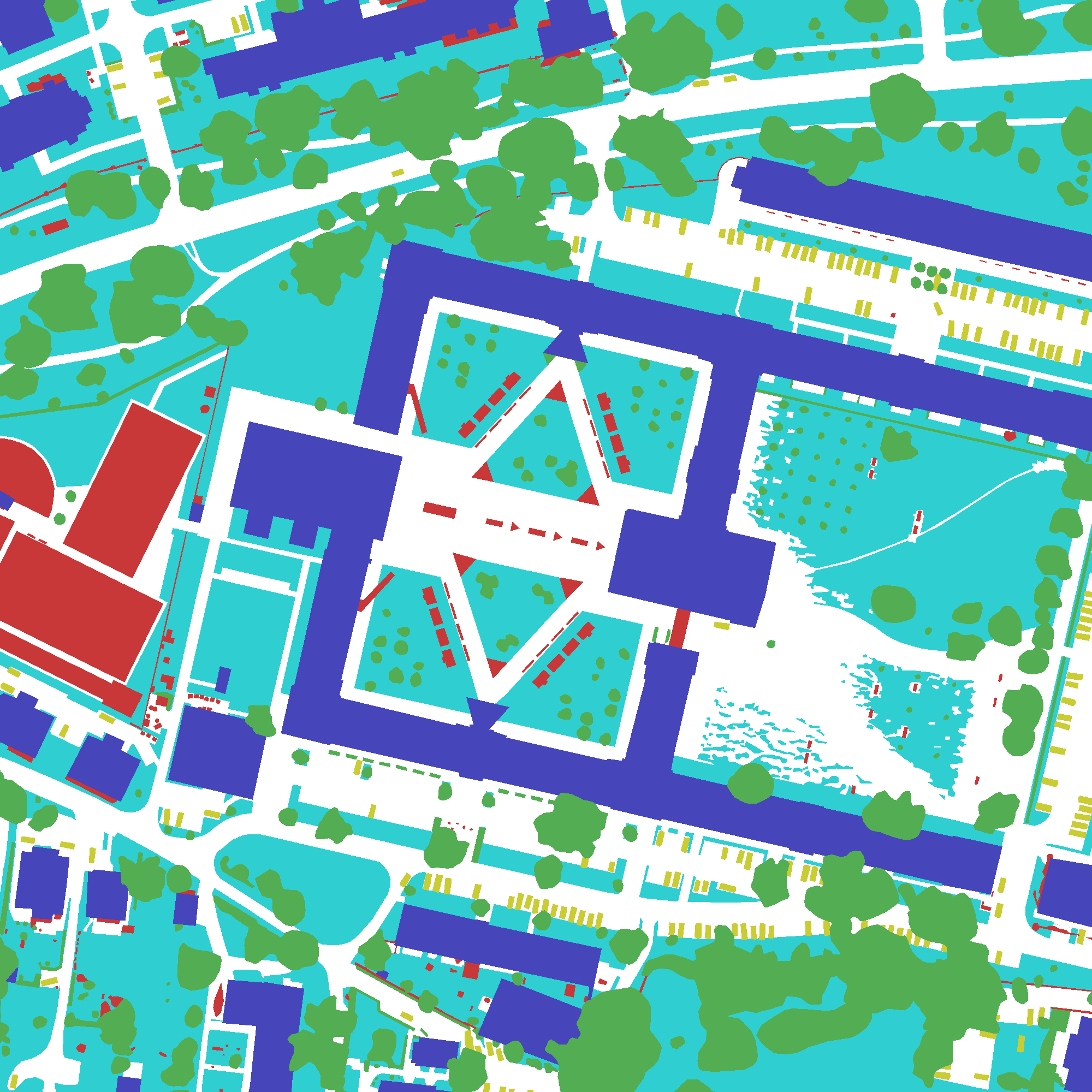}
    \caption{Ground truth}
    \end{subfigure}
    \begin{subfigure}{0.49\textwidth}
    \includegraphics[width=\linewidth]{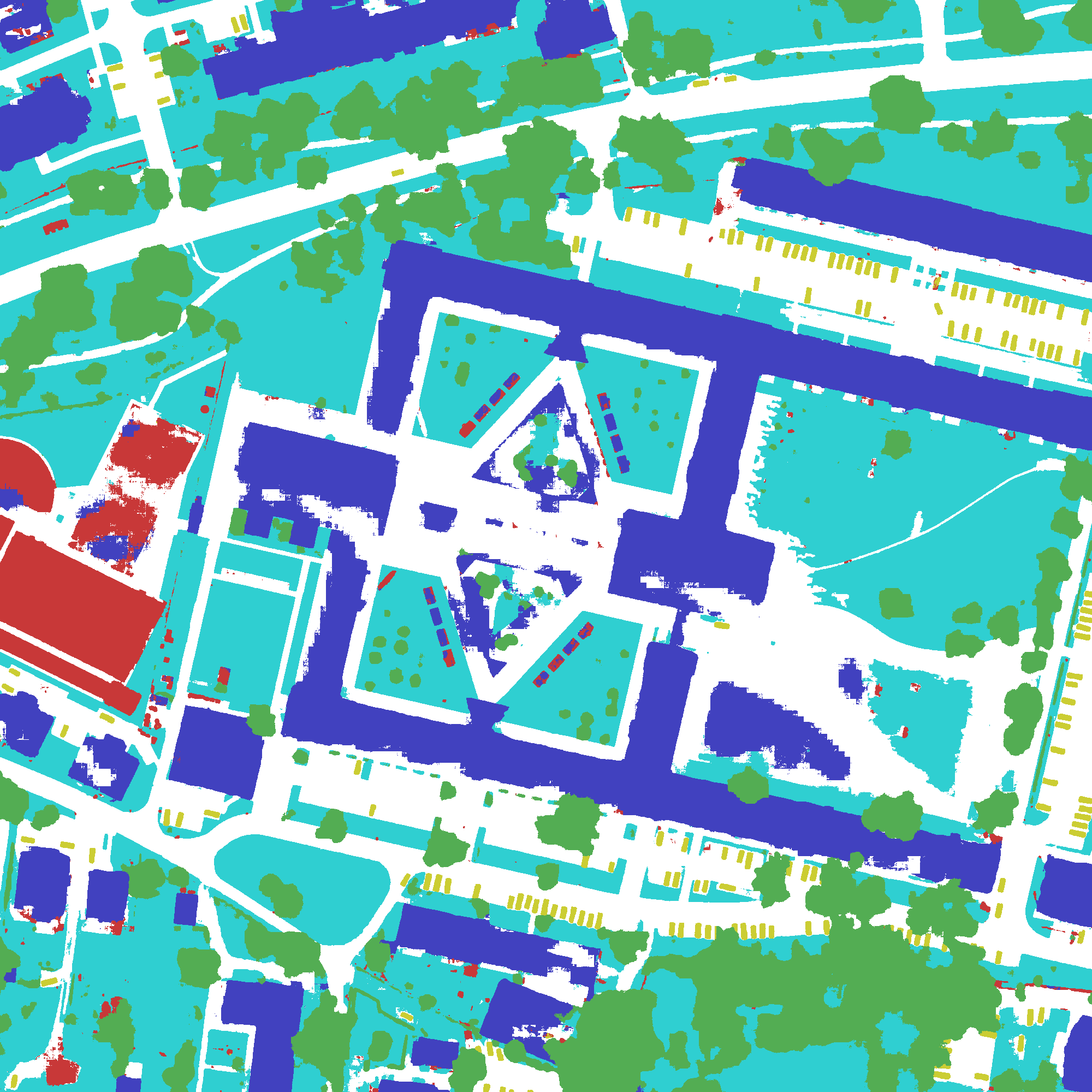}
    \caption{SegNet prediction}
    \end{subfigure}
    \begin{subfigure}{0.49\textwidth}
    \includegraphics[width=\linewidth]{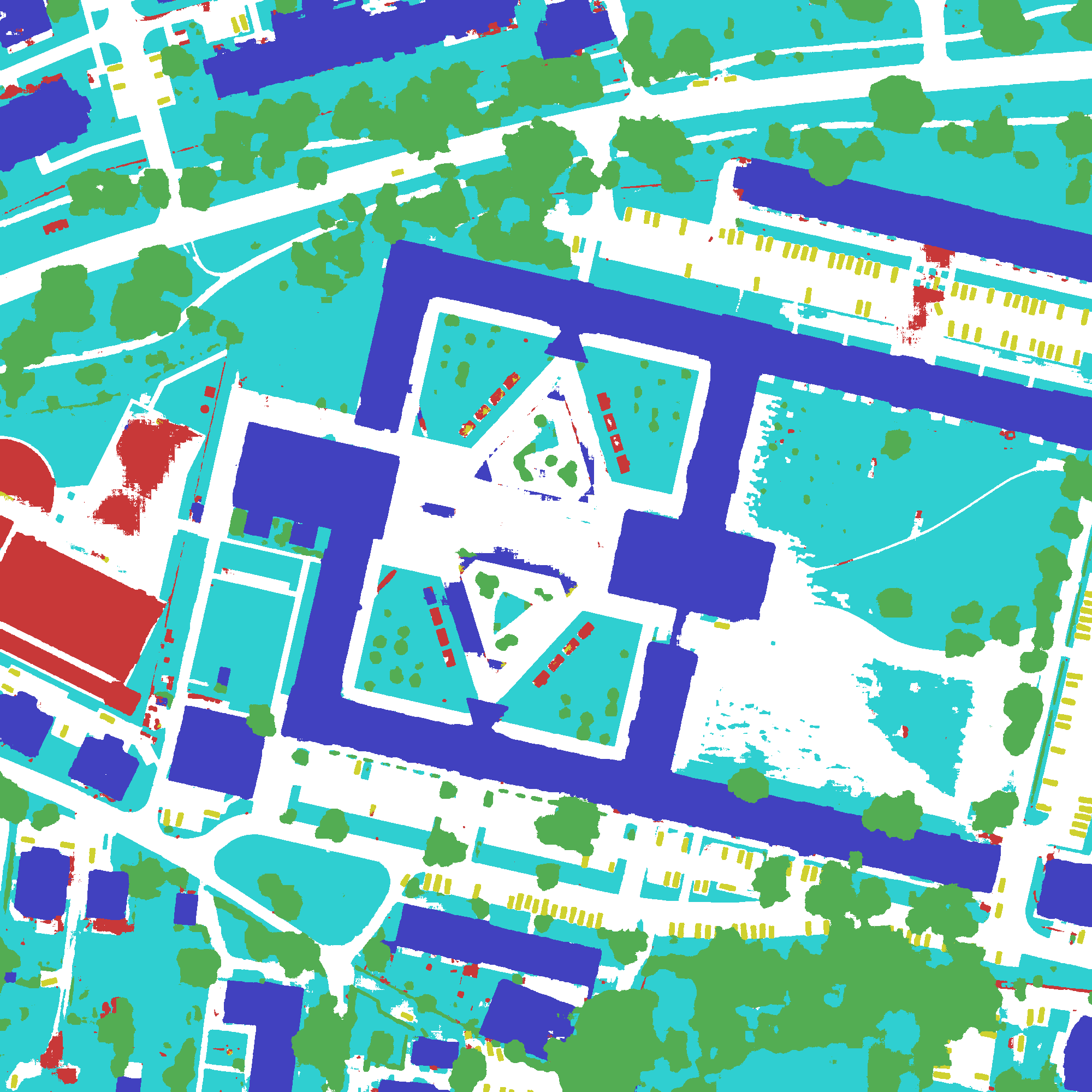}
    \caption{V-FuseNet prediction}
    \end{subfigure}
    \caption{Effect of the fusion strategy on an excerpt of the ISPRS Potsdam dataset. Confusion between impervious surfaces and buildings is significantly reduced thanks to the contribution of the nDSM in the V-FuseNet strategy.\\
(white: roads, {\color{blue!80!black} blue}: buildings, {\color{cyan!80!black} cyan}: low vegetation, {\color{green!80!black} green}:~trees, {\color{yellow!80!black} yellow}: cars)}
    \label{fig:potsdam_images}
\end{figure}

\cref{table:val_vaihingen} details the cross-validated results of our methods on the Vaihingen dataset.
We show the pixel-wise accuracy and the average F1 score over all classes. The F1 score over a class is defined by:
\begin{equation}
F1_{i} = 2~\frac{precision_{i} \times recall_{i}}{precision_{i} + recall_{i}}~,
\end{equation}
\begin{equation}
recall_i = \frac{tp_i}{C_i},~ precision_i = \frac{tp_i}{P_i}~,
\end{equation}
where $tp_i$ the number of true positives for class $i$, $C_i$ the number of pixels belonging to class $i$, and $P_i$ the number of pixels attributed to class $i$ by the model. As per the evaluation instructions from the challenge organizers, these metrics are computed after eroding the borders by a 3px radius circle and discarding those pixels.

\cref{table:msc_vaihingen} details the results of the multi-scale approach. ``No branch'' denotes the reference single-scale SegNet model. The first branch was added after the 4\textsuperscript{th} convolutional block of the decoder (downscale = 2), the second branch after the 3\textsuperscript{rd} (downscale = 4) and the third branch after the 2\textsuperscript{nd} (downscale = 8).

\cref{table:final_vaihingen} and \cref{table:final_potsdam} show the final results of our methods on the held-out test data from the Vaihingen and Potsdam datasets respectively.

\section{Discussion}
\label{S:discussion}

\begin{figure}[!tb]
\captionsetup[subfigure]{singlelinecheck=off,justification=centering}
\begin{subfigure}{0.5\textwidth}
  \captionsetup[subfigure]{singlelinecheck=off,justification=centering}
  \captionsetup[subfigure]{labelformat=empty}
	\begin{subfigure}{0.3\textwidth}
    	\includegraphics[width=\textwidth]{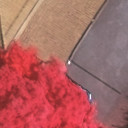}
        \caption*{IRRG image}
    \end{subfigure}
    \begin{subfigure}{0.3\textwidth}
    	\includegraphics[width=\textwidth]{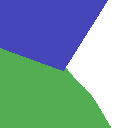}
        \caption*{Ground truth}
    \end{subfigure}
    \begin{subfigure}{0.3\textwidth}
    	\includegraphics[width=\textwidth]{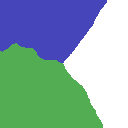}
        \caption*{SegNet prediction}
    \end{subfigure}
    \caption{SegNet can perform arguably better than the ground truth.}
    \label{fig:unprecise_transition}
\end{subfigure}
\begin{subfigure}{0.5\textwidth}
  \captionsetup[subfigure]{singlelinecheck=off,justification=centering}
  \captionsetup[subfigure]{labelformat=empty}
	\begin{subfigure}{0.3\textwidth}
    	\includegraphics[width=\textwidth]{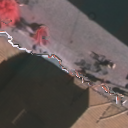}
        \caption*{IRRG image}
    \end{subfigure}
    \begin{subfigure}{0.3\textwidth}
    	\includegraphics[width=\textwidth]{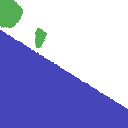}
        \caption*{Ground truth}
    \end{subfigure}
    \begin{subfigure}{0.3\textwidth}
    	\includegraphics[width=\textwidth]{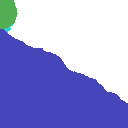}
        \caption*{SegNet prediction}
    \end{subfigure}
    \caption{SegNet sometimes overfits on geometrical aberrations.}
    \label{fig:geometric_dist}
\end{subfigure}
\caption{Disputable inconsistencies between our predictions and the ground truth.\\
(white: roads, {\color{blue!80!black} blue}: buildings, {\color{cyan!80!black} cyan}: low vegetation, {\color{green!80!black} green}:~trees, {\color{yellow!80!black} yellow}: cars)}
\end{figure}

\begin{figure}[!tb]
	\captionsetup[subfigure]{singlelinecheck=off,justification=centering}
    \captionsetup[subfigure]{labelformat=empty}
    \begin{subfigure}{0.24\textwidth}
    	\includegraphics[width=\textwidth]{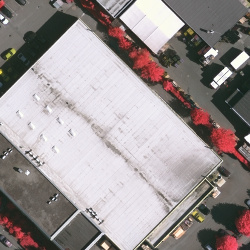}
        \caption{IRRG image}
    \end{subfigure}
    \begin{subfigure}{0.24\textwidth}
    	\includegraphics[width=\textwidth]{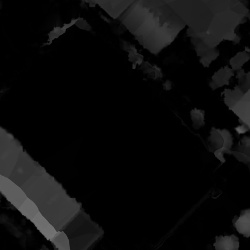}
        \caption{nDSM}
    \end{subfigure}
    \begin{subfigure}{0.24\textwidth}
    	\includegraphics[width=\textwidth]{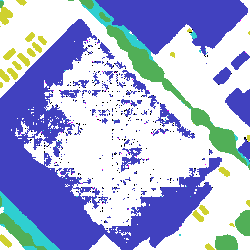}
        \caption{SegNet-RC}
    \end{subfigure}
    \begin{subfigure}{0.24\textwidth}
    	\includegraphics[width=\textwidth]{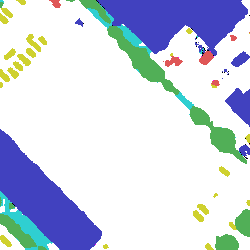}
        \caption{V-FuseNet}
    \end{subfigure}
	\caption{Errors in the Vaihingen nDSM are poorly handled by both fusion methods. Here, an entire building goes missing.}
    \label{fig:ndsm_fail}
\end{figure}

\subsection{Baselines and preliminary experiments}
As a baseline, we train standard SegNets and ResNets on the IRRG and composite versions of the Vaihingen and Potsdam datasets. These models are already competitive with the state-of-the-art as is, with a significant advance for the IRRG version. Especially, the car class has an average F1 score of $\simeq$~59.0\% on the composite images whereas it reaches $\simeq$ 85.0\% on the IRRG tiles. Nonetheless, we know that the composite tiles contain DSM information that could help on challenging frontiers such as roads/buildings and low vegetation/trees.

As illustrated in~\cref{table:val_vaihingen}, ResNet-34 performs slightly better in overall accuracy and obtains more stable results compared to SegNet. This is probably due to a better generalization capacity of ResNet that makes the model less subject to overfitting. Overall, ResNet and SegNet obtain similar results, with ResNet being more stable. However, ResNet requires significantly more memory compared to SegNet, especially when using the fusion schemes. Notably, we were not able to use the V-FuseNet scheme with ResNet-34 due to the memory limitation (12Gb) of our GPUs. Nonetheless, these results show that the investigated data fusion strategies can be applied to several flavors of Fully Convolutional Networks and that our findings should generalize to other base networks from the state-of-the-art.

\subsection{Effects of the multi-scale strategy}
The gain using the multi-scale approach is small, although it is virtually free as this only requires a few additional convolution parameters to extract downscaled maps from the lower layers. As could be expected, large structures such as roads and buildings benefit from the downscaled predictions, while cars are slightly less well detected in lower resolutions. We assume that vegetation is not structured and therefore the multi-scale approach does not help here, but instead increases the confusion between low and arboreal vegetation. Increasing the number of branches improves the overall classification but by a smaller margin each time, which is to be expected as the downscaled predictions become very coarse at 1:16 or 1:32 resolution.
Finally, although the quantitative improvements are low, a visual assessment of the inferred maps show that the qualitative improvement is non-negligible. As illustrated in~\cref{fig:vaihingen_images}, the multi-scale prediction regularizes and reduces the noise in the predictions. This makes it easier for subsequent human interpretation or post-processing, such as vectorization or shapefiles generation, especially on the man-made structures.

As a side effect of this investigation, our tests showed that the downscaled outputs were still quite accurate. For example, the prediction downscaled by a factor 8 was in average accuracy only 0.5\% below the full resolution prediction, with the difference mostly residing in ``car'' class. This is unsurprising as cars are usually $\simeq$~30px long in the full resolution tile and therefore cover only 3-4 pixels in the downscaled prediction, which makes them harder to see. Though, the good average accuracy of the downscaled outputs seems to indicate that the decoder from SegNet could be reduced to its first convolutional block without losing too much accuracy. This technique could be used to reduce the inference time when small objects are irrelevant while maintaining a good accuracy on the other classes.

\subsection{Effects of the fusion strategies}

\begin{figure}[!tb]
	\begin{subfigure}{\textwidth}
    	\captionsetup[subfigure]{singlelinecheck=off,justification=centering}
  		\captionsetup[subfigure]{labelformat=empty}
    	\begin{subfigure}{0.19\textwidth}
        	\includegraphics[width=\textwidth]{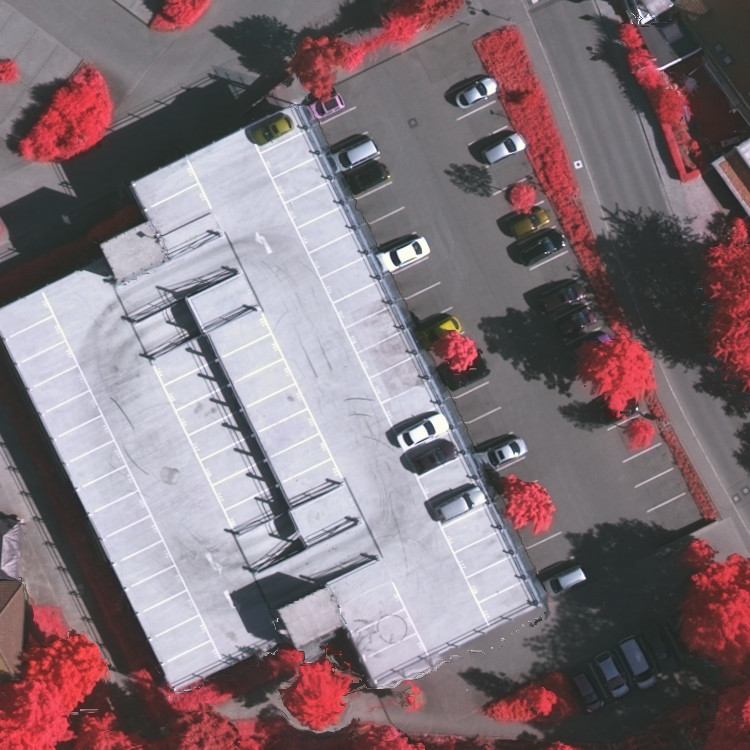}
      		\caption*{IRRG image}
        \end{subfigure}
        \begin{subfigure}{0.19\textwidth}
        	\includegraphics[width=\textwidth]{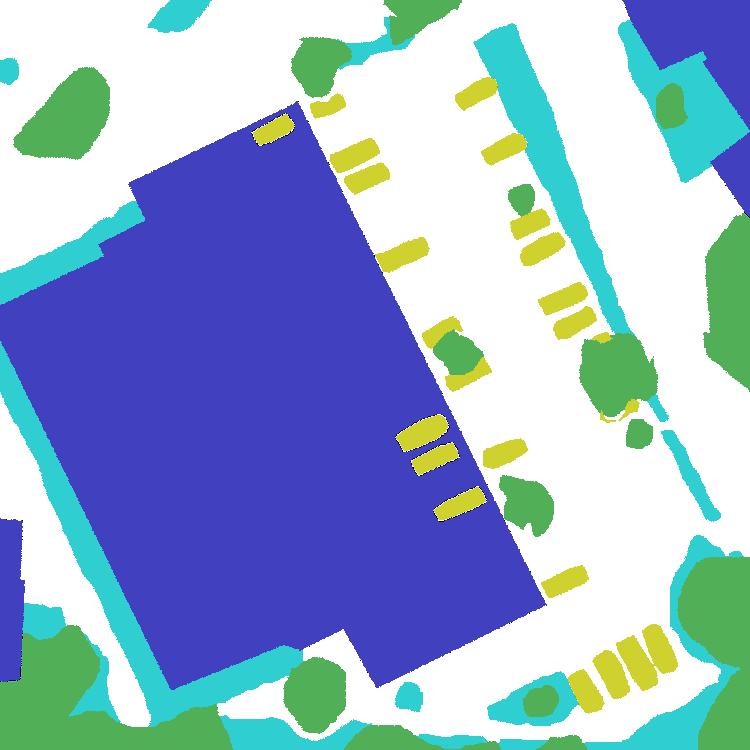}
        	\caption*{Ground truth}
        \end{subfigure}
        \begin{subfigure}{0.19\textwidth}
        	\includegraphics[width=\textwidth]{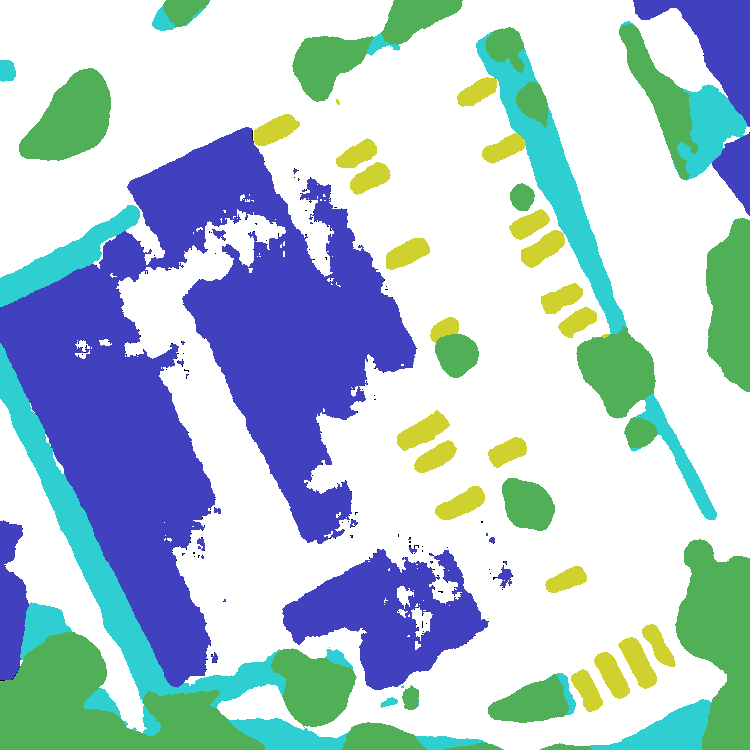}
        	\caption*{SegNet}
        \end{subfigure}
        \begin{subfigure}{0.19\textwidth}
        	\includegraphics[width=\textwidth]{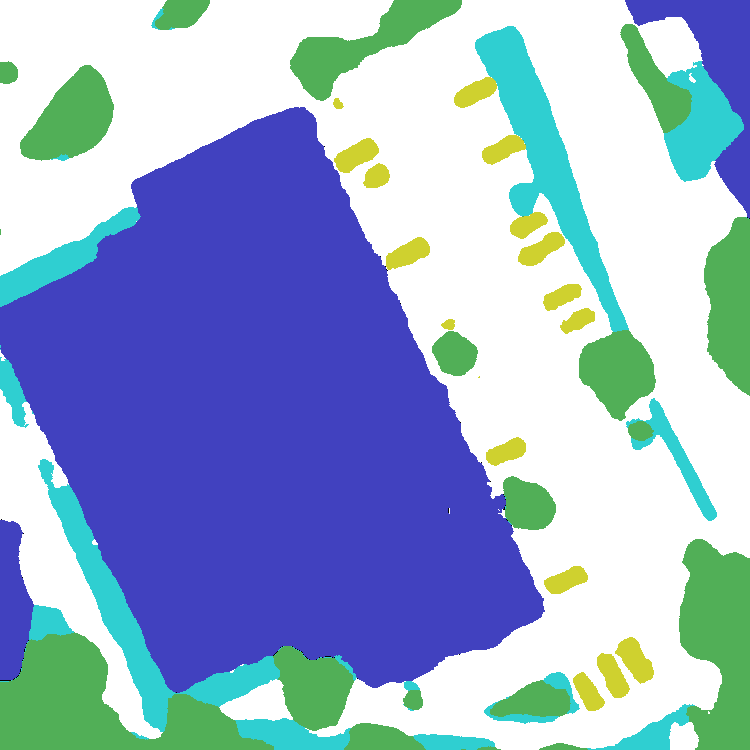}
        	\caption*{FuseNet}
        \end{subfigure}
        \begin{subfigure}{0.19\textwidth}
        	\includegraphics[width=\textwidth]{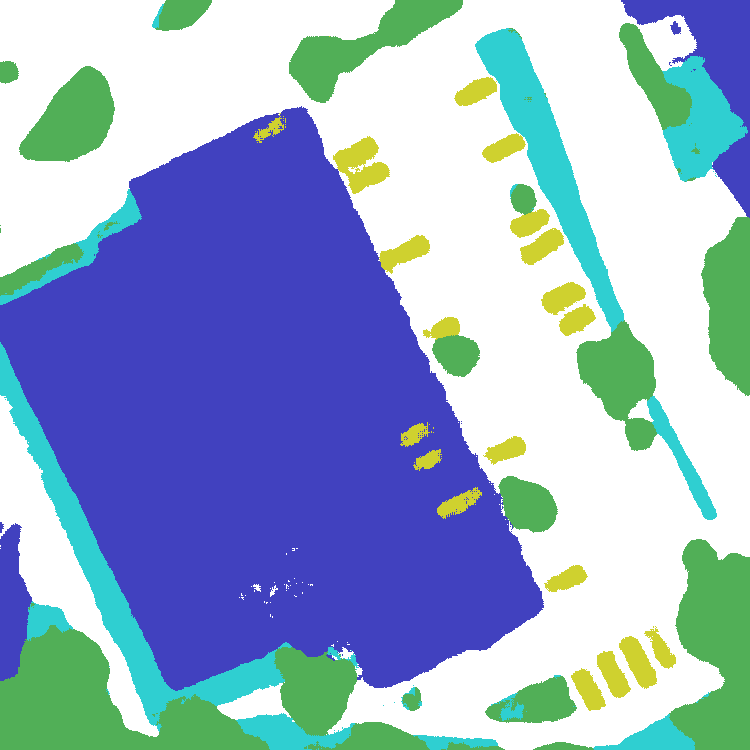}
        	\caption*{SegNet-RC}
        \end{subfigure}
        \caption{Predictions from various models on a patch of the Vaihingen dataset.}
        \label{fig:fusion_exemple1}
    \end{subfigure}
    	\begin{subfigure}{\textwidth}
    	\captionsetup[subfigure]{singlelinecheck=off,justification=centering}
  		\captionsetup[subfigure]{labelformat=empty}
    	\begin{subfigure}{0.19\textwidth}
        	\includegraphics[width=\textwidth]{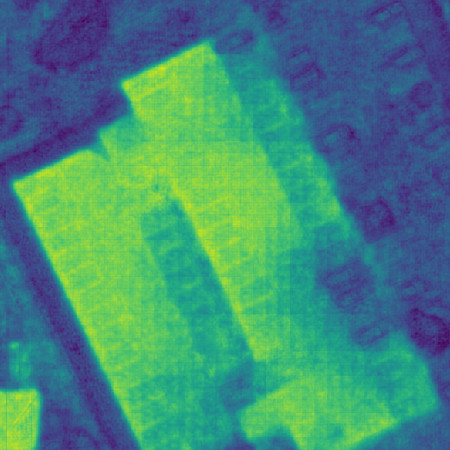}
      		\caption*{SegNet IRRG confidence (buildings)}
        \end{subfigure}
        \begin{subfigure}{0.19\textwidth}
        	\includegraphics[width=\textwidth]{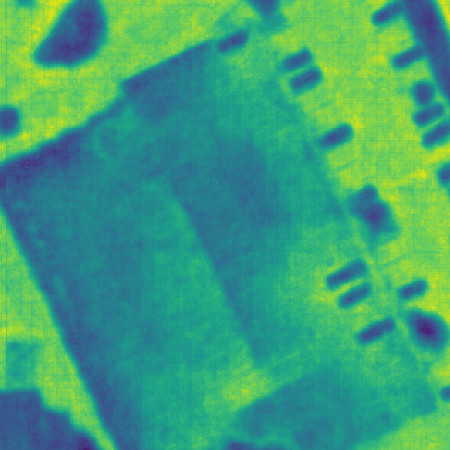}
        	\caption*{SegNet IRRG confidence (roads)}
        \end{subfigure}
        \begin{subfigure}{0.19\textwidth}
        	\includegraphics[width=\textwidth]{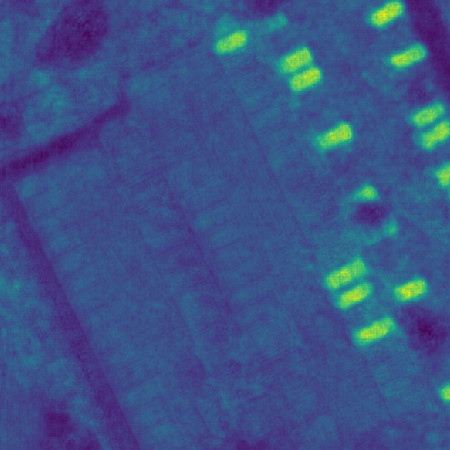}
        	\caption*{SegNet IRRG confidence (cars)}
        \end{subfigure}
        \begin{subfigure}{0.19\textwidth}
        	\includegraphics[width=\textwidth]{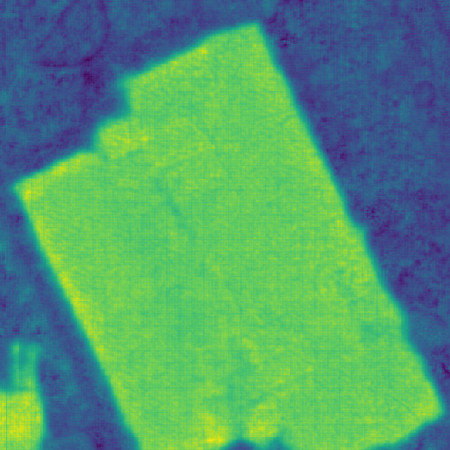}
        	\caption*{SegNet comp. confidence (buildings)}
        \end{subfigure}
        \begin{subfigure}{0.19\textwidth}
        	\includegraphics[width=\textwidth]{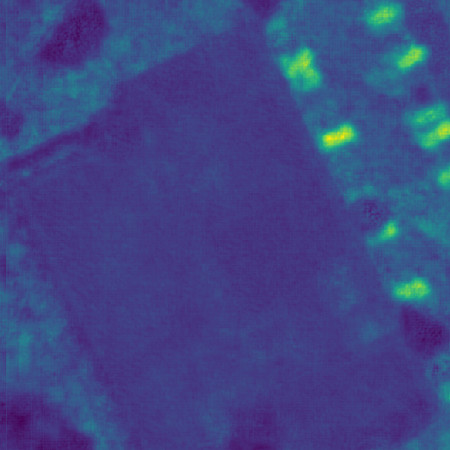}
        	\caption*{SegNet comp. confidence (cars)}
        \end{subfigure}
        \caption{SegNet confidence heat maps for various classes using several inputs.}
        \label{fig:confidence_vaihingen_fusion}
    \end{subfigure}
    \begin{subfigure}{\textwidth}
    	\captionsetup[subfigure]{singlelinecheck=off,justification=centering}
  		\captionsetup[subfigure]{labelformat=empty}
    	\begin{subfigure}{0.19\textwidth}
        	\includegraphics[width=\textwidth]{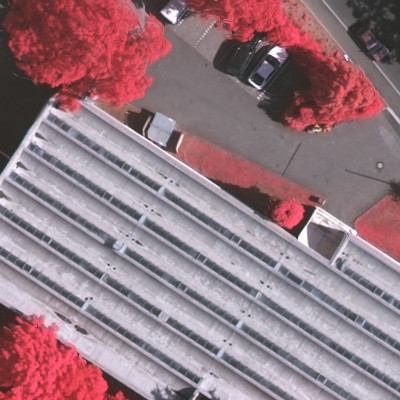}
      		\caption*{IRRG image}
        \end{subfigure}
        \begin{subfigure}{0.19\textwidth}
        	\includegraphics[width=\textwidth]{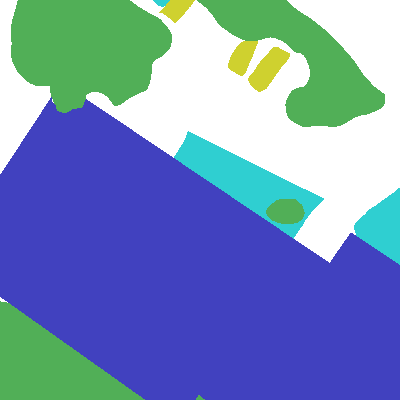}
        	\caption*{Ground truth}
        \end{subfigure}
        \begin{subfigure}{0.19\textwidth}
        	\includegraphics[width=\textwidth]{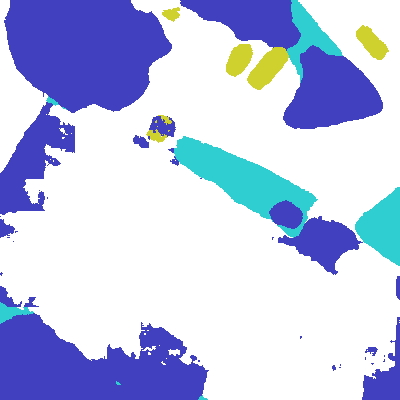}
        	\caption*{SegNet}
        \end{subfigure}
        \begin{subfigure}{0.19\textwidth}
        	\includegraphics[width=\textwidth]{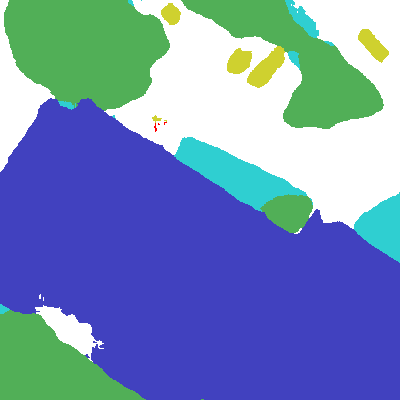}
        	\caption*{FuseNet}
        \end{subfigure}
        \begin{subfigure}{0.19\textwidth}
        	\includegraphics[width=\textwidth]{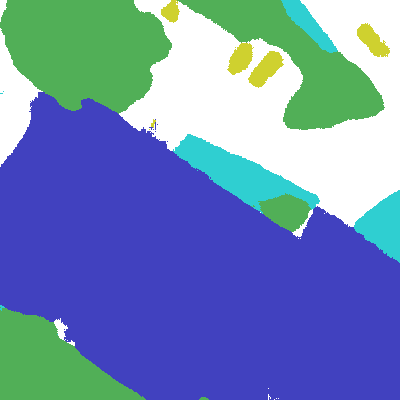}
        	\caption*{SegNet-RC}
        \end{subfigure}
        \caption{Predictions from various models on a patch of the Vaihingen dataset.}
        \label{fig:fusion_exemple2}
    \end{subfigure}
    
	\caption{Successful predictions using the fusion strategies.}
   	\label{fig:fusion_success}
\end{figure}

As expected, both fusion methods improve the classification accuracy on the two datasets, as illustrated in~\cref{fig:potsdam_images}. We show some examples of misclassified patches that are corrected using the fusion process in~\cref{fig:fusion_success}. In~\cref{fig:fusion_exemple1,fig:confidence_vaihingen_fusion}, SegNet is confused by the material of the building and the presence of cars. However, FuseNet uses the nDSM to decide that the structure is a building and ignore the cars, while the late fusion manages to mainly to recover the cars. This is similar to~\cref{fig:fusion_exemple2}, in which SegNet confuses the building with a road while FuseNet and the residual correction recover the information thanks to the nDSM both for the road and the trees in the top row.
One advantage of using the early fusion is that complementarity between the multiple modalities is leveraged more efficiently as it requires less parameters, yet achieves a better classification accuracy for all classes. At the opposite, late fusion with residual correction improves the overall accuracy at the price of less balanced predictions. Indeed, the increase mostly affects the ``building'' and ``impervious surface'' classes, while all the other F1 scores decrease slightly.

However, on the Potsdam dataset, the residual correction strategy slightly decreases the model accuracy. Indeed, the late fusion is mostly useful to combine strong predictions that are complementary. For example, as illustrated in~\cref{fig:confidence_vaihingen_fusion}, the composite SegNet has a strong confidence in its building prediction while the IRRG SegNet has a strong confidence in its cars predictions. Therefore, the residual correction is able to leverage those predictions and to fuse them to alleviate the uncertainty around the cars in the rooftop parking lot. This works well on Vaihingen as both the IRRG and composite sources achieve a global accuracy higher than 85\%. However, on Potsdam, the composite SegNet is less informative and achieves only 79\% accuracy, as the annotations are more precise and the dataset overall more challenging for a data source that relies only on Lidar and NDVI. Therefore, the residual correction fails to make the most of the two data sources. This analysis is comforted by the fact that, on the Vaihingen validation set, the residual correction achieves a better global accuracy with ResNets than with SegNets, thanks to the stronger ResNet-34 trained on the composite source.

Meanwhile, the FuseNet architecture learns a joint representation of the two data sources, but faces the same pitfall as the standard SegNet model : edge cases such as cars on rooftop parking lots disappear. However, the joint-features are significantly stronger and the decoder can perform a better classification using this multi-modal representation, therefore improving the global accuracy of the model.

In conclusion, the two fusion strategies can be used for different use cases. Late fusion by residual correction is more suited to combine several strong classifiers that are confident in their predictions, while the FuseNet early fusion scheme is more adapted for integrating weaker ancillary data into the main learning pipeline.

On the held-out testing set, the V-FuseNet strategy does not perform as well as expected. Its global accuracy is marginally under the original FuseNet model, although F1 scores on smaller and harder classes are improved, especially ``clutter'' which is improved from 49.3\% to 51.0\%. As the ``clutter'' class is ignored in the dataset metrics, this is not reflected in the final accuracy.

\subsection{Robustness to uncertainties and missing data}

As for all datasets, the ISPRS semantic labels in the ground truth suffer from some limitations. This can cause unfair mislabeling errors caused by missing objects in the ground truth or sharp transitions that do not reflect the true image (cf.~\cref{fig:unprecise_transition}).

However, even the raw data (optical and DSM) can be deceptive. Indeed, geometrical artifacts from the stitching process also impact negatively the segmentation, as our model overfits on those deformed pixels (cf.~\cref{fig:geometric_dist}).

Finally, due to limitations and noise in the Lidar point cloud, such as missing or aberrant points, the DSM and subsequently the nDSM present some artifacts. As reported in~\cite{marmanis_classification_2016}, some buildings vanish in the nDSM and the relevant pixels are falsely attributed a height of 0. This causes significant misclassification in the composite image that are poorly handled by both fusion methods, as illustrated in~\cref{fig:ndsm_fail}. \cite{marmanis_classification_2016} worked around this problem by manually correcting the nDSM, though this method does not scale to bigger datasets. Therefore, improving the method to be robust to impure data and artifacts could be helpful, e.g. by using hallucination networks~\cite{hoffman_learning_2016} to infer the missing modality as proposed in~\cite{kampffmeyer_semantic_2016}. We think that the V-FuseNet architecture could be adapted for such a purpose by using the virtual branch to encode missing data. Moreover, recent work on generative models might help alleviate overfitting and improve robustness by training on synthetic data, as proposed in~\cite{Xie_2017_ICCV}.

\section{Conclusion}

In this work, we investigate deep neural networks for semantic labeling of multi-modal very high-resolution urban remote sensing data. Especially, we show that fully convolutional networks are well-suited to the task and obtain excellent results. We present a simple deep supervision trick that extracts semantic maps at multiple resolutions, which helps training the network and improves the overall classification. Then, we extend our work to non-optical data by integrating digital surface model extracted from Lidar point clouds. We study two methods for multi-modal remote sensing data processing with deep networks: early fusion with FuseNet and late fusion using residual correction. We show that both methods can efficiently leverage the complementarity of the heterogeneous data, although on different use cases. While early fusion allows the network to learn stronger features, late fusion can recover errors on hard pixels that are missed by all the other models. We validated our findings on the ISPRS 2D Semantic Labeling datasets of Potsdam and Vaihingen, on which we obtained results competitive with the state-of-the-art.

\section*{Acknowledgements}
The Vaihingen dataset was provided by the German Society for Photogrammetry, Remote Sensing and Geoinformation (DGPF) \cite{cramer_dgpf_2010}: \url{http://www.ifp.uni-stuttgart.de/dgpf/DKEP-Allg.html}. The authors thank the ISPRS for making the Vaihingen and Potsdam datasets available and organizing the semantic labeling challenge. Nicolas Audebert's work is supported by the Total-ONERA research project NAOMI.




\section*{References}
\bibliography{GeoVision.bib}

\end{document}